\documentclass[10pt,journal,compsoc]{IEEEtran}

\usepackage{xspace}
\usepackage{url}
\usepackage{microtype}                 % use micro-typography (slightly more compact, better to read)
\PassOptionsToPackage{warn}{textcomp}  % to address font issues with \textrightarrow
\usepackage{textcomp}                  % use better special symbols
\usepackage{mathptmx}                  % use matching math font
\usepackage{times}                     % we use Times as the main font
\usepackage{cite}                      % needed to automatically sort the references
\usepackage{tabu}                      % only used for the table example
\usepackage{booktabs}                  % only used for the table example
\usepackage{amsmath}
\usepackage{amssymb}
\usepackage{mathrsfs}
\usepackage[bottom]{footmisc}
\usepackage{graphics}
\usepackage{multirow}
\usepackage{multicol}
\usepackage{color}
\usepackage{transparent}
\usepackage[export]{adjustbox}
\usepackage{dblfloatfix}
\usepackage{xcolor}
\usepackage{subcaption}
\usepackage{enumitem}% http://ctan.org/pkg/enumitem
\usepackage[font=small]{caption}
\usepackage[pscoord]{eso-pic}% The zero point of the coordinate systemis the lower left corner of the page (the default).

\newcommand{\placetextbox}[3]{% \placetextbox{<horizontal pos>}{<vertical pos>}{<stuff>}
  \setbox0=\hbox{#3}% Put <stuff> in a box
  \AddToShipoutPictureFG*{% Add <stuff> to current page foreground
    \put(\LenToUnit{#1\paperwidth},\LenToUnit{#2\paperheight}){\vtop{{\null}\makebox[0pt][c]{#3}}}%
  }%
}%
\newcommand{\toolname}{DPVis\xspace}
\newcommand{\static}{Static Variables View\xspace}
\newcommand{\dist}{State Summary Panel\xspace}
\newcommand{\heatmap}{Feature Matrix\xspace}
\newcommand{\detail}{Feature Distribution\xspace}

\newcommand{\transit}{State Transitions Panel\xspace}
\newcommand{\timepathway}{Pathway by Time Unit\xspace}
\newcommand{\obspathway}{Pathway over Observation\xspace}

\newcommand{\pathwayfall}{Pathway Waterfall\xspace}
\newcommand{\chord}{State Transition Chord Diagram\xspace}
\newcommand{\freq}{State Transition Pattern List\xspace}
\newcommand{\subjecttimeline}{Panel for Patterns and Outcomes\xspace}
\newcommand{\subjectlist}{Subject List View\xspace}
\newcommand{\kernel}{Dual Kernel Densities View\xspace}
\newcommand{\subgroup}{Subgroup Builder\xspace}
\newcommand{\query}{State Sequence Finder\xspace}
\newcommand{\cohort}{Subgroup List\xspace}

\newcommand{\alignment}{Alignment Handle\xspace}

\usepackage{graphicx}
\usepackage{tikz}

\newcommand*\annotatedFigureBoxCustom[8]{ (#1) rectangle (#2);\node at (#4) [fill=#6,thick,shape=circle,draw=#7,inner sep=2pt,font=\sffamily,text=#8,opacity=0.0] {\textbf{#3}};}
\newcommand*\annotatedFigureBox[4]{\annotatedFigureBoxCustom{#1}{#2}{#3}{#4}{white}{white}{black}{black}}

\newenvironment {annotatedFigure}[1]{\centering\begin{tikzpicture}
\node[anchor=south west,inner sep=0,opacity=0.0] (image) at (0,0) { #1};\begin{scope}[x={(image.south east)},y={(image.north west)}]}{\end{scope}\end{tikzpicture}}
\begin{document}
%% Paper title.
\title{\toolname: Visual Analytics with Hidden Markov Models for Disease Progression Pathways}
\placetextbox{0.5}{.99}{\small\textcolor{blue}{To appear in IEEE Transactions on Visualization and Computer Graphics. DOI: 10.1109/TVCG.2020.2985689. Details: \url{https://bckwon.com/publication/dpvis}}}%
%%%%%%%%%%%%%%%%%%%%%%%%%%%%%%%%%%%%%%%%%%%%%%%%%%%%%%%%%%%%%%%%%%%%%%%

%% This is how authors are specified in the journal style

%% indicate IEEE Member or Student Member in form indicated below
\author{Bum Chul Kwon, Vibha Anand, Kristen A. Severson, Soumya Ghosh,\\ Zhaonan Sun, Brigitte I. Frohnert, Markus Lundgren, and Kenney Ng
\IEEEcompsocitemizethanks{\IEEEcompsocthanksitem Bum Chul Kwon, Vibha Anand, Kristen A. Severson, Soumya Ghosh, Zhaonan Sun, and Kenney Ng are with IBM Research. BCK, VA, and KN are also with the T1DI Study Group. E-mail: \{bumchul.kwon, anand, kristen.severson,  ghoshso, zsun, kenney.ng\}@us.ibm.com.\protect\\
\IEEEcompsocthanksitem  Brigitte I. Frohnert is with University of Colorado Denver and the T1DI Study Group. E-mail: brigitte.frohnert@ucdenver.edu.\protect\\
\IEEEcompsocthanksitem Markus Lundgren is with Department of Clinical Sciences Malm\"o, Lund University, Sweden and the T1DI Study Group. E-mail: markus.lundgren@med.lu.se.}% <-this % stops an unwanted space
% \thanks{Manuscript received XX, 2019; revised XX, 2019; Accepted XX, 2019}
\thanks{Manuscript received XX, 2019; revised XX, 2020; Accepted XX, 2020}
}

% %other entries to be set up for journal
% \shortauthortitle{Kwon \MakeLowercase{\textit{et al.}}: \toolname: Visual Analytics with Hidden Markov Models for Disease Progression Pathways}

\IEEEtitleabstractindextext{%
\begin{abstract}

Clinical researchers use disease progression models to understand patient status and characterize progression patterns from longitudinal health records. 
One approach for disease progression modeling is to describe patient status using a small number of states that represent distinctive distributions over a set of observed measures.
Hidden Markov models (HMMs) and its variants are a class of models that both discover these states and make inferences of health states for patients.
Despite the advantages of using the algorithms for discovering interesting patterns, it still remains challenging for medical experts to interpret model outputs, understand complex modeling parameters, and clinically make sense of the patterns.
To tackle these problems, we conducted a design study with clinical scientists, statisticians, and visualization experts, with the goal to investigate disease progression pathways of chronic diseases, namely type 1 diabetes (T1D), Huntington's disease, Parkinson's disease, and chronic obstructive pulmonary disease (COPD).
As a result, we introduce \toolname which seamlessly integrates model parameters and outcomes of HMMs into interpretable and interactive visualizations.
In this study, we demonstrate that \toolname is successful in evaluating disease progression models, visually summarizing disease states, interactively exploring disease progression patterns, and building, analyzing, and comparing clinically relevant patient subgroups.
\end{abstract}

%% Keywords that describe your work. Will show as 'Index Terms' in journal
%% please capitalize first letter and insert punctuation after last keyword
\begin{IEEEkeywords}
Disease Progression, Hidden Markov Model, State Space Model, Diabetes, Huntington's, Parkinson's, Interpretability
\end{IEEEkeywords}
}

\maketitle
\IEEEdisplaynontitleabstractindextext
\IEEEpeerreviewmaketitle

\IEEEraisesectionheading{\section{Introduction}\label{sec:intro}}
\IEEEPARstart{C}{linical} researchers want to understand the progression of diseases.
Improved understanding of disease progression can enable a variety of clinical tasks, including patient management, cohort selection, and drug discovery. 
To achieve this goal, clinical researchers often employ quantitative models, termed `disease progression models' (DPMs), to characterize the course of disease progression from patients' data contained in longitudinal health records.
Using the model, researchers aim to gain insights about how characteristics of patients interact with the evolution of disease. 
This knowledge can ultimately lead to early detection of diseases and precision care for each patient at appropriate points in time.

However, understanding disease progression trajectories is a non-trivial task. 
First, modeling progression of diseases is challenging because it is often manifested by multiple symptoms over time, which makes it difficult to summarize progression patterns.
Second, clinical studies usually observe the symptoms of multiple subjects over discrete, irregular time points, which make modeling the patterns even more difficult.
Third, clinical researchers not only want to summarize disease progression patterns of subjects, but also analyze the relationship between their trajectories and health outcomes.
Likewise, understanding disease progression patterns involve modeling the distinct state transitions from complex, longitudinal patient records and analyzing the association between learned transition patterns and various measures (e.g., genetic profiles) in order to derive clinically meaningful insights.

Research on type 1 diabetes is an illustrative example for the importance of DPMs.
In the current understanding, type 1 diabetes development can be divided into three stages: i) Stage-1: normal blood sugar; ii) Stage-2: abnormal blood sugar; iii) Stage-3: clinical diagnosis of type 1 diabetes~\cite{insel_staging_2015}.
However, the three stages assume two or more autoantibodies, and it is not yet established how individuals progress from no autoantibodies to multiple autoantibodies.
There may be differences in the progression speed between pre-diabetic individuals depending on the combinations of autoantibodies that they possess at different time points.
Furthermore, some individuals with different genetic predispositions may follow different trajectories of the evolution of autoantibodies.
To gain more insights, researchers want to discover transition patterns between high-level `states' that are manifested by onsets of autoantibodies, to explore subjects' transition patterns between the states, and find association between various genetic variables and trajectories.

In this study, we conducted a design study with clinical researchers so that we build a tool that helps them to visually explore disease progression patterns.
Throughout the design study, we derived our users' task hierarchy and explored design space to support the tasks using interactive visualization techniques.
We chose Hidden Markov Models (HMMs), over other candidates because they can infer discrete latent states and transitions between the inferred states from time-varying multivariate data.
In particular, they adequately handle longitudinal medical data from clinical studies, which are often incomplete, missing, and measured at irregular time points.
In essence, a trained HMM infers a state (output) per patient visit based on a set of observed measures (input) at the visit.
For example, T1D researchers can train a HMM with three autoantibodies variables (input) and use the trained model to infer states (output) of patients' records over time. 
If they choose a model with \emph{K} states, then each visit will be assigned to one of \emph{K} states. 
The number of states is a hyperparameter, which is typically chosen based on cross validation or prior knowledge. Typically the number of visits is far larger than the number of states thus causing multiple visits to be assigned to the same state. The characteristics of the state --- a statistical summary of all visits assigned to the state, is learned automatically from the data.
Using the states inferred (labeled) by the HMM, researchers can understand the evolution of autoantibodies of multiple patients by inspecting which states subjects tend to go through at different ages and which state transitions occur more frequently.
As a result of our design study, we developed a visual analytics system called \toolname that incorporates HMMs with interactive visualizations.
Our case study demonstrates the usefulness of \toolname for disease progression analysis.
The main contributions of the paper are summarized as following:

\begin{enumerate}
\item We developed \toolname that incorporates HMMs with interactive visualizations for exploring disease progression patterns from longitudinal health records.
\item Our design study examines tasks of clinical researchers who aim to understand disease progression stages and provides various ways to visualize the statistical outcomes of HMMs for supporting these tasks.
\item We report a usage scenario and users' experiences that demonstrates the usefulness of \toolname for domain experts to interpret HMMs in a transparent manner, to detect interesting disease progression patterns, and to derive clinical insights.
\end{enumerate}

\section{Related Work}
\label{sec:related}

In this section, we review disease progression studies in the field of medicine, modeling techniques in statistics and machine learning, and visual analytic approaches in the visualization domain.

\subsection{Disease Progression Research in Clinical Studies}
\label{sec:clinical}
Well-designed longitudinal studies play an important role in clinical research because they provide rich information for tracking the progression of disease by collecting repeated measures from individuals of a target cohort over a long period of time.   
Such studies are designed and conducted with many present and future goals, such as understanding the etiology or pathogenesis of a disease.
One of the most well-known longitudinal studies is the Framingham Heart Study, which has brought great insights into heart disease (e.g.,~\cite{Levy1990, Hubert1983}).
In particular, many studies call for research into applying suitable statistical modeling techniques that can explain the mechanism of disease progressions.
For example, the Parkinson Progression Marker Initiative (PPMI)~\cite{marek2011} is an international observational multi-center study of Parkinson's Disease. 
PPMI established protocols that were adopted at 21 clinical sites to provide a platform for researchers to gain common access to the data, which includes clinical, socio-demographic, and imaging variables as well as bio-specimens.
In case of T1D, researchers seek to understand the diverse pathways from a healthy state to the onset of disease for both individual patients and the overall population~\cite{ziegler_accelerated_2011}.
The Type 1 Data Intelligence (T1DI) study group was established to accelerate disease progression research by combining decades of data from birth cohort studies conducted at various institutions around the world: 1) Diabetes Autoimmunity Study in the Young (DAISY)~\cite{rewers_newborn_1996}; 2) Diabetes Prediction in Sk{\aa}ne (DiPiS)~\cite{jonsdottir_childhood_2018}; 3) Diabetes Prediction and Prevention (DIPP)~\cite{nejentsev_population-based_1999}; and 4) Diabetes Evaluation in Washington State (DEW-IT)~\cite{wion_population-wide_2003}.
We conducted a design study with clinical researchers and statisticians who have been involved in the research groups above.

\subsection{Disease Progression Modeling}
\label{sec:dpm}

{\em Disease Progression Models} (DPM)~\cite{mould2012models} include a broad class of models that quantitatively predict disease status over time. 
DPM can be categorized by algorithm into systems biology, data-driven, and semi-mechanistic models~\cite{Cook2016}. 
There are many different approaches to model disease progression, including path models~\cite{VogelsteinEtAl1988}, tree based models~\cite{DesperEtAl1999}, hierarchical latent variable models~\cite{Schulam2015}, Gaussian process models~\cite{Lorezi2017}, and Bayesian networks~\cite{BeerenwinkelEtAl2004}. 
In this study, we have chosen to focus on Hidden Markov Models (HMMs), a class of probabilistic generative models commonly used to model data with longitudinal dependencies. 
In an HMM, the target disease is represented by a fixed number of typical disease states, its progression is characterized by a Markov process, and the observed measures are the manifestations of the underlying disease states. 
By characterizing the progression of disease as a series of transitions between typical disease states, HMMs provide an inherently interpretable summary of the progression. Moreover, HMMs exhibit several desirable properties that make them well suited for disease progression modeling.  
First, they are probabilistic in nature and are able to represent uncertainties in clinical data organically. 
Second, they are able to handle missing data, a common issue in observational studies. Third, users can apply their domain knowledge into modeling disease progression dynamics by setting various constraints. For instance, users can prohibit a model from stepping backward in diseased states when they investigate chronic diseases, where the damage caused by the disease is irreversible and affected areas gradually lose their functions, and backward-enabled progression, where patients can recover from the diseases. Fourth, HMMs are unsupervised models that do not require a particular outcome task. Fifth, they provide interpretability by a small number of `states' that are associated with multiple observed variables. 
Examples of variants of HMMs applied to disease progression include \cite{Sukkar2012, WangEtAl2014_KDD, Sun2019_JAMIA, LiuLiEtAl2015_NIPS,JacksonEtAl2003_JRSS}. 
HMMs are described further in Section~\ref{sec:HMM}.
Despite the advantages, it still remains challenging for clinical scientists to use the HMM-based models for exploring disease progression patterns without visual aids. To use the disease progression model, experts need (i) to understand the output probabilities of diverse features for individual states, (ii) to investigate differences in state transition patterns between patients, and (iii) to associate such patterns with patient outcome variables (e.g., onset, death, discharge).

\subsection{Visualizations for Sequential Event Analysis and Disease Progression Analysis}
\label{sec:dpmvis}

In this section, we review previous studies that show inspiring examples to tackle our problem.
Prior studies have investigated various visualization methods to represent temporal event sequences.
Shneiderman and Plaisant~\cite{shneiderman_interactive_2019} summarize the challenges of analyzing temporal sequence data.
Several techniques were proposed to integrate sequence mining algorithms for visualizing the summary of event sequences, including LifeLines~\cite{plaisant_lifelines_1998},  LifeLines2~\cite{wang_temporal_2009}, EventFlow~\cite{monroe_temporal_2013}, and LifeFlow~\cite{wongsuphasawat_lifeflow:_2011}, which allow users to visually align and explore various patterns of multiple event sequences.
Frequence~\cite{perer_frequence:_2014}, Care Pathway Explorer~\cite{perer_mining_2015}, and Peekquence~\cite{kwon_peekquence_2016}  visualize frequent event sequences mined from the SPAM algorithm~\cite{ayres_sequential_2002}.
Outflow~\cite{wongsuphasawat_exploring_2012} and Sequence Synopsis~\cite{chen_sequence_2018} allow users to interactively explore multiple pathways in event sequences.
These techniques allow users to steer the algorithm with interaction so that they can find meaningful summaries of event sequences.
However, these studies often deal with pre-defined event types that are often independent of feature values associated with each event, which is different from statistically learned states that are associated with various features in HMMs. Thus, the visualization approaches for pre-defined event types fall short in revealing the relationship between the HMM states and the associated variables.

Other techniques integrate automated results with user-defined queries or criteria.
TimeStitch~\cite{polack_timestitch_2015}, Choronodes~\cite{polack_chronodes_2018}, and Coquito~\cite{krause_supporting_2016} extract frequent mining patterns using algorithms like PrefixSpan~\cite{pei_prefixspan_2001}, and then visualizes the sequences so that users can interactively provide relevant feedback to refine the search.
Eventpad~\cite{cappers_exploring_2018} and $(s|qu)eries$~\cite{zgraggen_s_2015} allow users to build search queries using the progressive visual analytics (PVA) paradigm~\cite{stolper_progressive_2014}; the idea was further extended and implemented as the PPMT tool~\cite{raveneau_progressive_2018}.
EventThread~\cite{guo_eventthread:_2018} visualizes clusters of event sequences using tensor analysis.
Liu et al.~\cite{liu_patterns_2017}, MAQUI~\cite{law_maqui:_2019}, Cadence~\cite{gotz2019visual}, and VASABI~\cite{nguyen2019vasabi} allow users to recursively explore hierarchical patterns in event sequences.
ET$^{2}$~\cite{guo_visual_2019},  StageMap~\cite{chen_stagemap:_2018}, and Mathisen and GrØnbæk~\cite{mathisen_clear_2017} introduce a composite event sequence to aggregate patterns.
CoreFlow~\cite{liu_coreflow:_2017} and Guo et al.~\cite{guo_visualizing_2019} visualize branching alternative paths with uncertainties.
IDMVis~\cite{zhang_idmvis_2019} allows users to fold and align records to derive event sequence patterns.
Though query-based techniques are useful and inspiring, it requires more sophisticated operations to build queries based on various characteristics of HMMs, such as posterior distributions (uncertainties) over observed variables and time ranges of event occurrences.

Previous studies also investigate visual analytic methods to explore and discover patterns from longitudinal data in clinical studies and electronic medical records.
DecisionFlow~\cite{gotz_decisionflow_2014} allows users to explore multidimensional event sequences using multiple, coordinated visual analytics systems.
Researchers also investigate methods to analyze heterogeneous cohorts using interactive visualizations~\cite{angelelli2014interactive,klemm2014interactive}.
PhenoBlocks~\cite{glueck_phenoblocks_2016}, PhenoStacks~\cite{glueck_phenostacks:_2017}, and PhenoLines~\cite{glueck_phenolines_2018} allow users to explore and compare cohorts based on various measures.
RetainVis, together with RetainEX, allows users to understand how individual visits and categorical variables contribute to making diagnostic risk predictions~\cite{kwon_retainvis_2019}.
Clustervision~\cite{kwon18cluster} help users to find informative cohorts of patients based on their common diseases, treatments, diagnostic measures, and comorbidities.
Bernard et al. developed and evaluated static dashboard network that can help users to observe longitudinal changes of multiple patients~\cite{bernard_using_2019}.
Other studies help users make sense of time-series health data, such as Stroscope~\cite{cho_stroscope_2014}, TimeSpan~\cite{loorak_timespan_2016}, Marai et al.~\cite{marai_precision_2019}, and RegressionExplorer~\cite{dingen_regressionexplorer_2019}.
Previous studies show inspiring examples to coordinate multiple views for event sequence visualization. \toolname needs to tailor these approaches for showing the similarities and differences in state transition patterns between a variety of different subgroups of patients that can be defined from each view.

Though the previous approaches provide inspiring techniques, there is no unified approach that integrates state-space models such as HMMs for exploring disease progression patterns from longitudinal observational data.
Our literature review shows that we need to adapt the approaches for our data, model, and tasks. The following section describes the adaptation process and result.

\section{Design Study Method}
\label{sec:method}
This section describes our design study by introducing characteristics of domain experts, data, HMMs, and tasks and requirements that experts intend to achieve through visualizations.

\subsection{Target Users: Clinical Researchers}
\label{sec:user_characteristics}

We joined a collaborative project established in 2017 by JDRF and their academic partners for computational modeling of T1D: the T1DI (Type 1 Data Intelligence) study group~\cite{hu_ai_2019}.
Other research groups with progression modeling studies for Parkinson's and Huntington's disease were invited to collaborate and provide feedback.
Our primary target users are clinical researchers, whose goal is to investigate disease progression patterns in observational data collected from clinical studies.

\subsection{Domain Goals and Tasks}
\label{sec:domain_questions}

In the \textit{T1DI} study group, we organized more than ten conference calls and four workshops between October, 2018 and October, 2019 to discuss disease progression trajectories of type 1 diabetes.
The participants were clinical researchers who conduct and investigate observational studies on children with type 1 diabetes.
In the meetings, we focused on identifying clinically meaningful research questions and hypotheses domain experts would like to ask on the observational data through iterative discussion.
Authors, as participants of the meetings, consolidated the questions and hypotheses into three high-level research goals below (\textbf{G1--3}).

A goal of clinical researchers is to develop optimal treatments tailored for individuals by estimating their disease progression patterns precisely.
Thus, they want to be able to discover distinct states of patients that are associated with clinical diagnosis.
Early and accurate detection of presymptomatic progression signals can lead to early intervention and precision medicine for target patients.
To achieve this goal, clinical researchers want to characterize and summarize the disease progression patterns by a set of biomarkers, lab tests, and other measures in a data-driven, unsupervised manner.
Ultimately, clinical researchers hope to gain clinically useful insights about the disease progression patterns through subjects' longitudinal data collected from observational studies.

Clinical researchers want to understand disease progression patterns using observation data.
Understanding disease progression patterns can be divided into three analytic goals.
First, researchers want to \emph{explain disease progression} by summarizing the evolution of selected variables (\textbf{G1}).
Explaining disease progression includes the discovery of distinctive states that are characterized by the selected variables.
For example, researchers want to describe the evolution of autoantibodies that subjects possess and lose before they get diagnosed with type 1 diabetes.
In doing so, they want to find the unique combination of multiple autoantibodies multiple subjects show before diagnosis.
Second, they want to \emph{discover heterogeneous trajectory groups} (\textbf{G2}).
They want to find distinctive trajectory groups of patients whose records show common patterns in terms of the evolution of selected variables.
For example, clinical researchers want to identify typical progression patterns in terms of the order of autoantibodies subjects gain before they are diagnosed with type 1 diabetes.
They also want to discover how many subjects follow a specific progression pattern.
Lastly, they want to \emph{find associations between specific trajectories and variables} (\textbf{G3}). 
For example, they may hypothesize that subjects with a specific genetic profile may show rapid progression by gaining multiple autoantibodies simultaneously at an early age.

\subsection{Observation Data and Hidden Markov Models}
\label{sec:HMM}

\begin{figure}[tb]
    \centering
    \includegraphics[width=.5\textwidth, frame, trim={.0cm .25cm .0cm 7.0cm},clip]{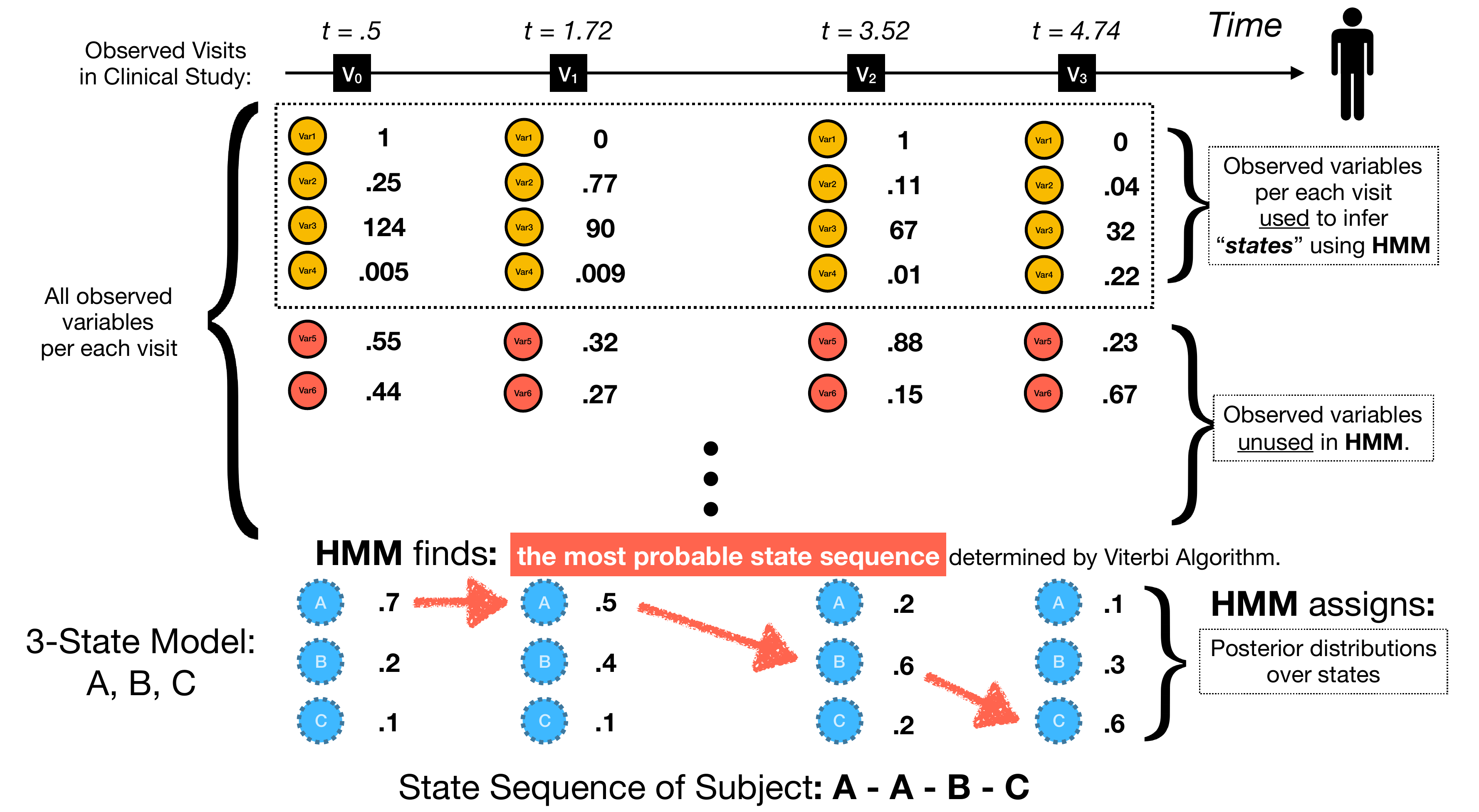}
    \caption{HMMs learn disease states based on observed attribute values chosen by users for training the model. The trained model predicts the most probable state sequences for a patient and assigns posterior probabilities for each visit.}
    \label{fig:cthmm}
    \vspace{-.5cm}
\end{figure}

To expose distinct states of disease progression from the evolution of observed variables, we use HMMs to model sequences of hidden states over time on longitudinal observational data.
As Fig.~\ref{fig:cthmm} shows, a patient's record includes multiple visits (black squares) spread over different time points (t).
In each visit, a patient goes through multiple procedures, which generate observed variables (circles).
Clinical researchers choose a set of variables (yellow circles) among them in order to discover hidden states using HMMs; we call the variables as output variables.
Prior to training, in addition to the choice of output variables, they decide how many hidden states they are going to infer and what constraints they want to impose on progression dynamics.
To determine the number of states, researchers conduct cross-validation analysis to measure the model fitness and evaluate the usefulness of the model instances based on clinical knowledge by exploring exposed patterns.
Constraints can be interpreted as any restriction in state transitions.
If we consider a 3-state model as Fig.~\ref{fig:cthmm} shows, we can set a constraint on transition probabilities so that a patient can only advance to the next step at a time without going backward or skipping.
Researchers often choose the restrictions for more easily interpretable trajectory groups or based on prior knowledge about diseases (e.g., no-backward restriction for chronic diseases).
Once the training is done, researchers used the trained model to infer hidden states for every visit.
As part of inference, the model generates posterior probabilities (blue dotted circles) over hidden states discovered.
Clinical researchers can investigate the state sequence of subjects (pink arrows).
The HMM outcomes, hidden-state labels and posterior probabilities over them for every visit, combined with observed variables and other demographic variables are great sources to understand granular disease progression patterns of multiple subjects.

\subsection{Visual Disease Progression Analysis Tasks}
\label{sec:tasks_requirements}

In this section, we translate domain tasks into visualization tasks by considering the outputs of HMMs discussed earlier.
The first three tasks (T1, T2, T3) are drawn from the three domain tasks. 
Then, T4 and T5 were added to address researchers' need to investigate individual patients' records and to analyze complex subgroups, respectively.
The tasks are used to guide our design decisions on various features of \toolname.

\begin{description}[style=unboxed,leftmargin=0cm,nosep]
\item[T1: Characteristics of states:] To understand the distinct states discovered by HMM-based models, clinical researchers want to view the distribution of multiple variables per each state. Understanding what each state means is also important for users to evaluate whether the discovered states clinically make sense. Understanding characteristics of states includes gaining an overview of distinct states, understanding the evolution of variables for each state, and understanding differences between the states.
\item[T2: State transition patterns:] Clinical researchers want to discover the state transition patterns from multiple subjects. For each state transition pattern, clinical researchers want to investigate details, such as the number of subjects who follow the pattern and when subjects make transition from one state to another state. Investigation of state transition patterns involve discovery of distinct state sequence patterns, understanding the heterogeneity of the patterns with respect to the time of transitions, and comparison between trajectories by the number of subjects, speed of progression, and how many states are involved.
\item[T3: Relationship between trajectories and variables:] Clinical researchers want to find the association between state transition patterns and variables like health outcomes and static variables. For example, they want to investigate whether certain trajectories correlate with the onset of the target disease. They also want to find association between other inherent characteristics of subjects, such as genetic profiles, and trajectories. Thus, they want to view distribution of static/outcome variables for trajectory groups, view the onsets of disease for trajectory groups, and compare the static/outcome variables between trajectory groups.
\item[T4: Subject details:] Clinical researchers want to investigate details by viewing state transitions, health outcomes, and other variables of a single subject. They also want to find similarities and differences between multiple subjects with respect to observation counts, times, and state transition patterns.
\item[T5: Subgroup management:] Clinical researchers build and refine subgroups based on transition patterns, health outcomes, and subject profiles. Then, they compare similarities and differences among them. Subgroups are important outcomes of their research, which need to be retained and transferable to other colleagues. 
\end{description}

\begin{figure*}[t!]
  \centering
  \begin{annotatedFigure}
	{\includegraphics[width=\linewidth, frame, trim={.0cm 0.1cm .0cm 1.8cm},clip]{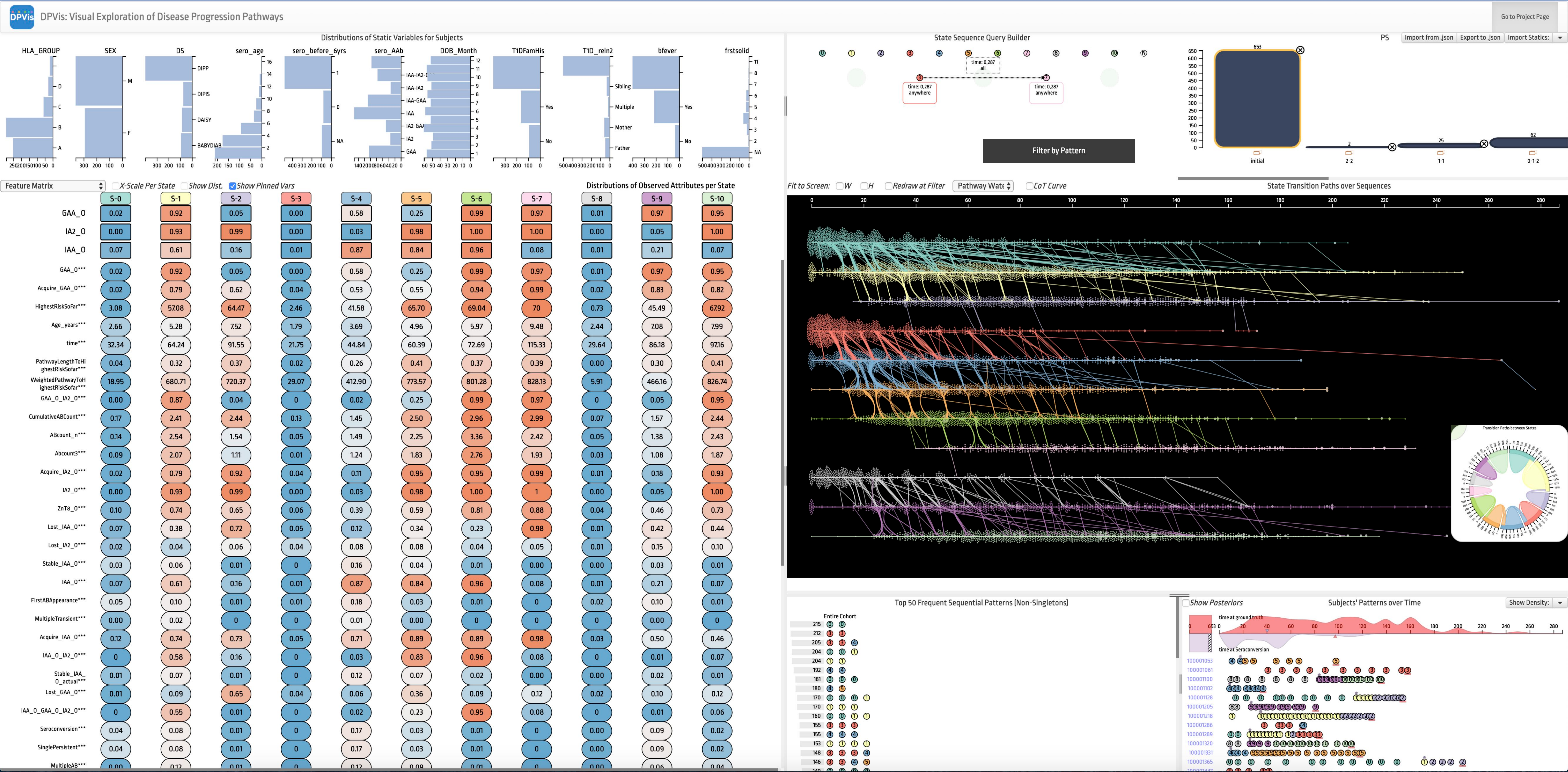}}
	\annotatedFigureBox{0.050,0.82}{0.2896,0.94}{A}{0.020,0.95}%tl
	\annotatedFigureBox{0.050,0.82}{0.089,0.715}{B}{0.020,0.76}%tl
	\annotatedFigureBox{0.502,0.62}{0.547,0.715}{C}{0.502,0.76}%tl
	\annotatedFigureBox{0.502,0.1493}{0.702,0.2693}{D}{0.502,0.2693}%tl
	\annotatedFigureBox{0.751,0.1493}{1.151,0.2693}{E}{0.751,0.2693}%tl
	\annotatedFigureBox{0.502,0.82}{.547,0.94}{F}{0.502,0.95}%tl
	\annotatedFigureBox{0.751,0.82}{1.151,0.94}{G}{0.751,0.95}%tl
\end{annotatedFigure}
\caption{Given a dataset of patients over time and the corresponding disease state assignment per visit as determined by an HMM, \toolname creates the following views: (A) \static shows distribution over static variables; (B) \heatmap summarizes states, discovered by HMM, with means of variables; (C) \pathwayfall shows state transition patterns; (D) \freq show frequently occurring state transition patterns; (E) \subjecttimeline shows subjects' visits in detail; (F) \query enables users to build and refine cohorts; (G) \cohort allows users to create and refine subgroups.}
	\label{fig:teaser}
	\vspace{-.25in}
\end{figure*}

\section{Design of \toolname}
\label{sec:design}
Based on the derived tasks, we designed and assessed multiple views and operations and integrated them into a visual analytics application. 
In this section, we introduce the design of \toolname and how each view and interaction feature supports the user's analytic tasks.
\toolname consists of seven view panels as Fig.~\ref{fig:teaser} shows, and each plays a different role for one or more of five tasks we described in Section~\ref{sec:tasks_requirements}.
We built \toolname using Python for the backend and Javascript for the frontend. We used Flask, Jinja, and Django for the web framework, and primarily used D3.js, jQuery, and Lodash to implement UI components, visualizations, and interactive features.
In this section, we use type 1 diabetes as a running example to describe insights and clinical explanations from observational data.
We explore disease progression patterns of 559 subjects, who were diagnosed with type 1 diabetes at the end of their observation, from birth cohort studies, using an 11-state HMM model.
Find more details about the study, model, and data in Section~\ref{sec:casestudy}.

\subsection{\dist}

\dist (Fig.~\ref{fig:teaser}~(B)) contains two views, \heatmap and \detail, which characterize states with respect to observed attributes.
Using each view, users can gain an overview and details about states by viewing the distribution of observed attributes per state (\textbf{T1}). 
Furthermore, users can make sense of differences between states that are discovered by HMMs.

\subsubsection{\heatmap}

\heatmap summarizes how each state is manifested by different observed attributes (\textbf{T1}).
\heatmap shows a matrix layout where a row represents an observed attribute and a column represents a state;
each cell in this layout represents the mean of the attribute (row) values of the visits that were inferred as the corresponding state (column).
The diverging color map is used to map the range of minimum-maximum values of each measure to the blue-red scale across the states.
Using this view, users can understand what each state (column) means by viewing the distribution over multiple attributes (row).
In Fig.~\ref{fig:teaser}~(B), the leftmost column is 'State 0', which has cell values in mostly blue.
This indicates that the 'State 0' indicates states, where subjects have low values for the majority of observed attributes.
Thus, researchers can note that the state is likely to be before the onset of autoimmunity.
In particular, we repeated and fixed attributes (rows) that are used to infer states at the top so that users can quickly refer to them while exploring state transitions in Fig.~\ref{fig:teaser}. 
At first, we considered a heatmap-style visualization where no space exists between rows and between columns. 
While the heatmap approach can be useful to detect correlation, users had difficulty in isolating column-wise and row-wise patterns due to the absence of borders.
This can be potentially resolved by adding highlights when users hover particular state. 
However, users quickly glance at this view to be reminded about state characteristics overview, so it is cumbersome to hover over the view every time.
Therefore, we decided to use the button-like approach for \heatmap.

\subsubsection{\detail}

\begin{figure}[htb]
    \centering
    \includegraphics[width=.485\textwidth, frame, trim={3.5cm 0.0cm .0cm 0.0cm},clip]{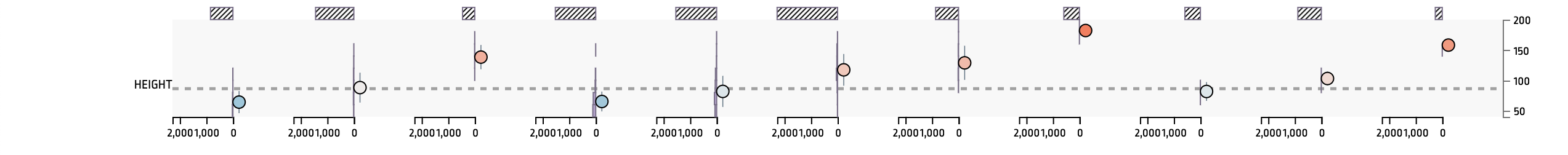}
    \caption{\detail shows mean, standard deviation, and histogram of an attribute `HEIGHT' across 11 states.}
    \label{fig:dist}
    \vspace{-.25cm}
\end{figure}

The states are probabilistically inferred by HMMs, so users often need to understand standard deviation as well as mean for each observed attribute.
Greater uncertainties may indicate that it is possible to have sub-states within a single state, so users can expect more granular, diverse patterns with the state.
In \detail, users can view mean, standard deviation, and histogram of each observed attribute per state in a single cell, replacing colored cells in \heatmap. 
In addition, users can check the number of missing values per cell with a shaded bar above each cell.
Fig.~\ref{fig:dist} shows the distribution of an attribute `HEIGHT' across 11 states.
First, there are many missing values for the `HEIGHT' measure, which can be observed by large, shaded bars above each state.
We can conjecture that such growth variables (e.g., height, weight, BMI) were not regularly measured through the observational study.
With the limitations in mind, the mean and standard deviation shows the steady increase of HEIGHT within each of three notable state sequences, namely 0--2, 3--7, and 8--10.
Within each block, the mean height increases.
Also, the last state of each block shows narrow intervals (standard deviation) and dense histogram, which may indicate that those last states are likely to be `sink states' where subjects get diagnosed with Type 1 Diabetes.
We implemented several ways to customize the view further.
First, users can switch and toggle back to fit the scale for each cell.
If users choose a consistent axis, users can compare the number of visits assigned to states. If users switch to fit the scale for each state, users can clearly observe distribution among individual state, which could be visually hidden on a common scale.
Second, users can simplify the view by hiding bars. Users revealed that it is easier to compare mean and variance between selected subgroups and all subjects when there are no distracting bars.

\begin{figure*}[tb]
    \captionsetup[subfigure]{labelformat=empty}
    \centering
    \begin{subfigure}{.49\textwidth}
        \centering
        \includegraphics[width=\textwidth, trim={.0cm 2.0cm .0cm 1.25cm},clip]{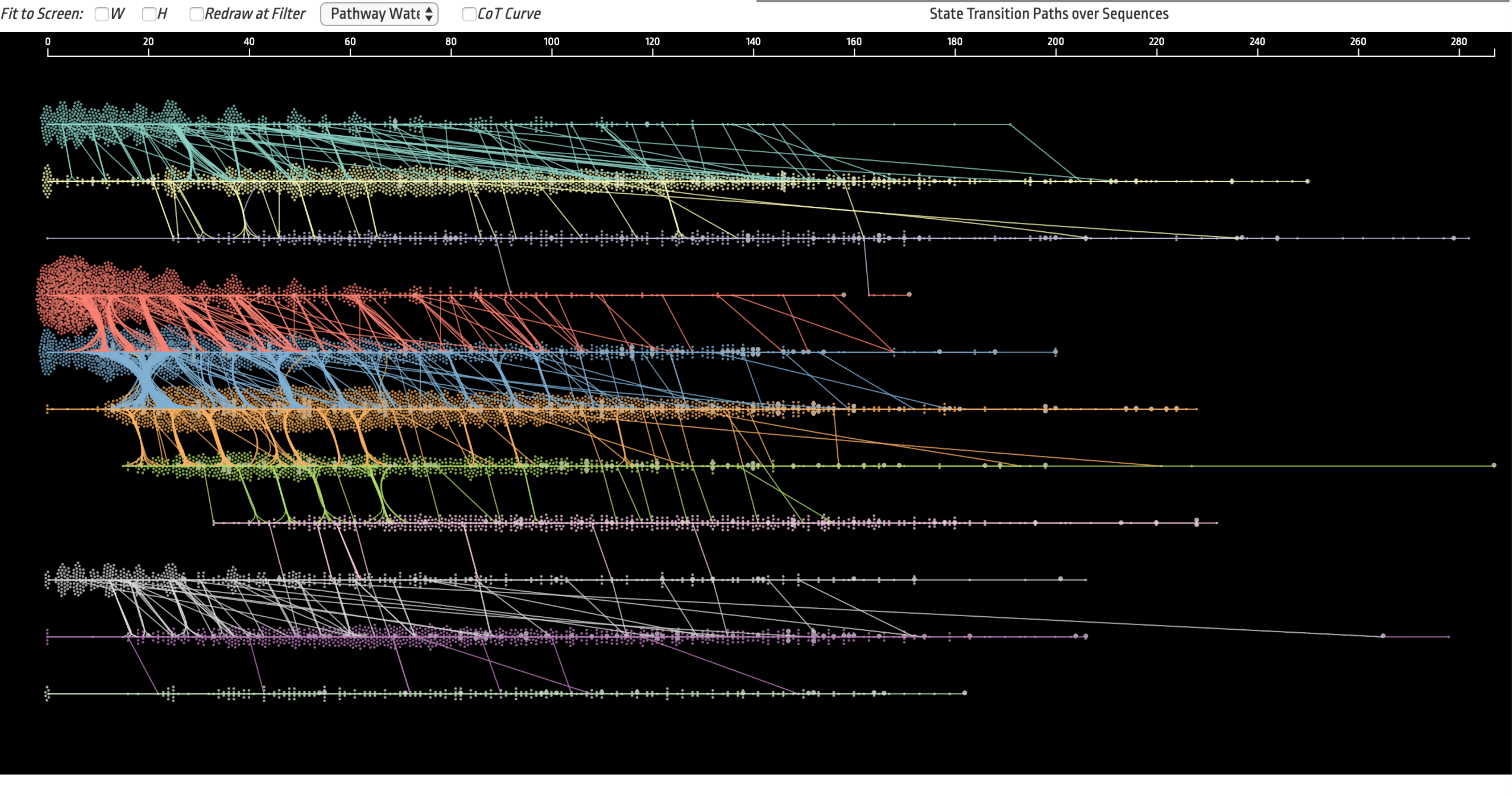}
        \vspace{-.5cm}
        \label{fig:pathwayfall1}
    \end{subfigure}
    \begin{subfigure}{.49\textwidth}
        \centering
        \includegraphics[width=\textwidth, trim={.0cm 2.0cm .0cm 1.25cm},clip]{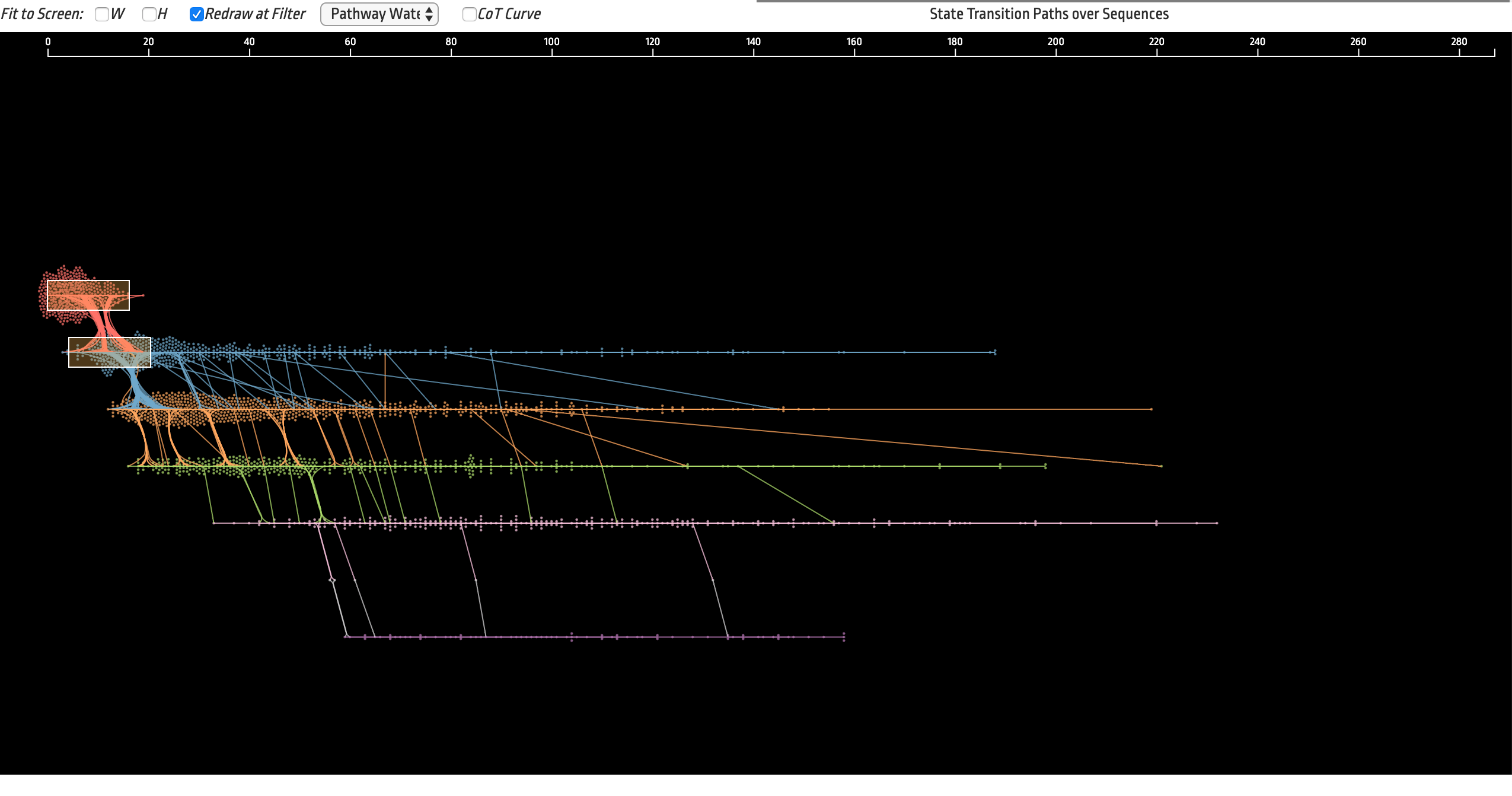}
        \vspace{-.5cm}
        \label{fig:pathwayfall2}
    \end{subfigure}
    \caption{\pathwayfall shows state transition pathways using parallel beeswarm plots and trajectory lines tied with force edge bundling: dot (visit), y-axis \& color: state, x-axis: age. Vertical paths show transitions between states. \textbf{Left}: state transition patterns of all subjects over; \textbf{Right}: users filtered by state transition from state 3 to 4 before 20 months of age.}
    \vspace{-.5cm}
    \label{fig:pathwayfall}
\end{figure*}

\begin{figure*}[b!]
    \vspace{-.5cm}
    \captionsetup[subfigure]{labelformat=empty}
    \centering
    \begin{subfigure}{.33\textwidth}
        \centering
        \includegraphics[width=\textwidth, trim={.0cm 0.0cm 8.0cm 1.25cm},clip]{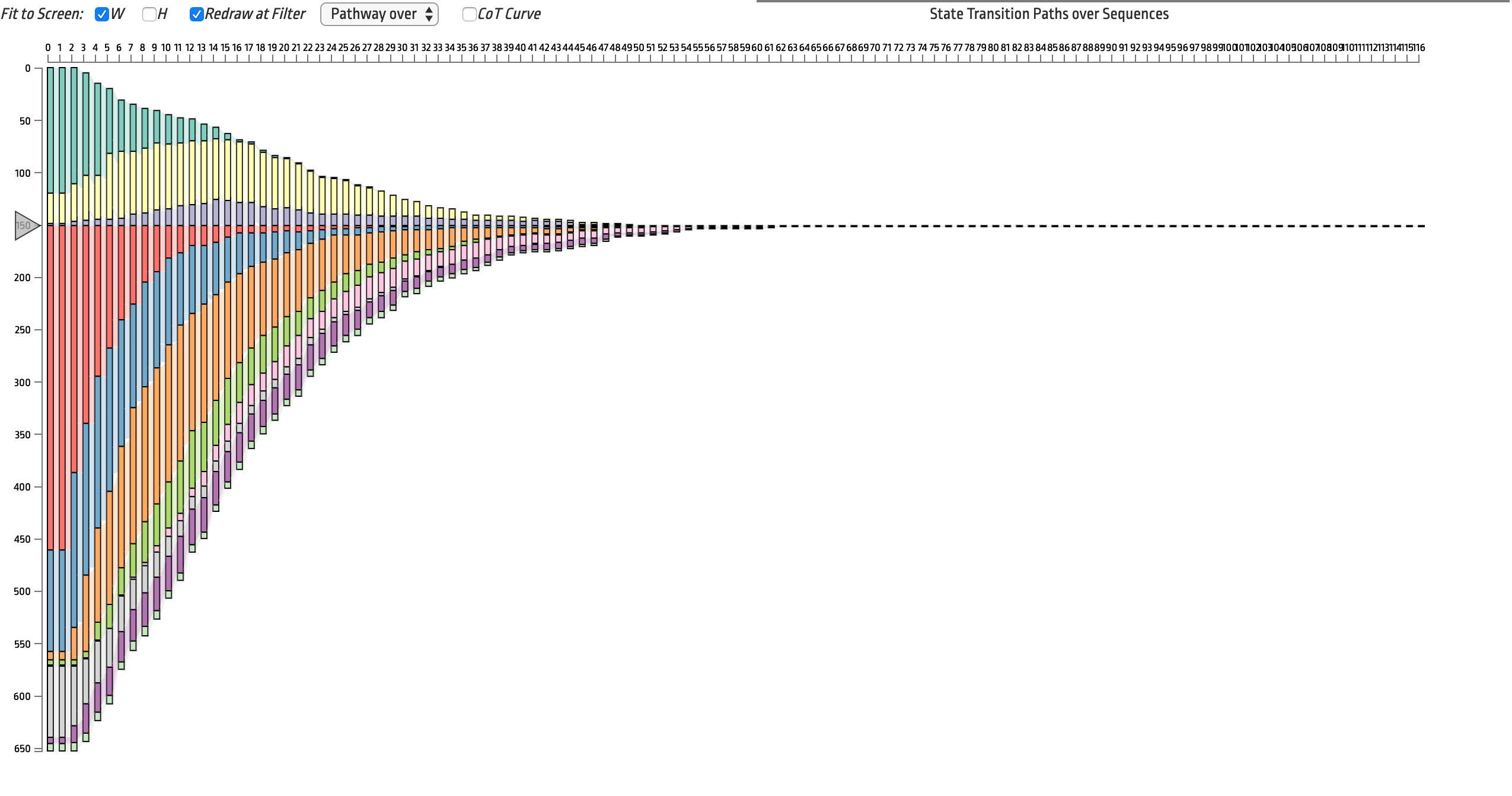}
        \vspace{-.5cm}
        \label{fig:obspathway1}
    \end{subfigure}
    \begin{subfigure}{.33\textwidth}
        \centering
        \includegraphics[width=\textwidth, trim={.0cm 0.0cm 8.0cm 1.25cm},clip]{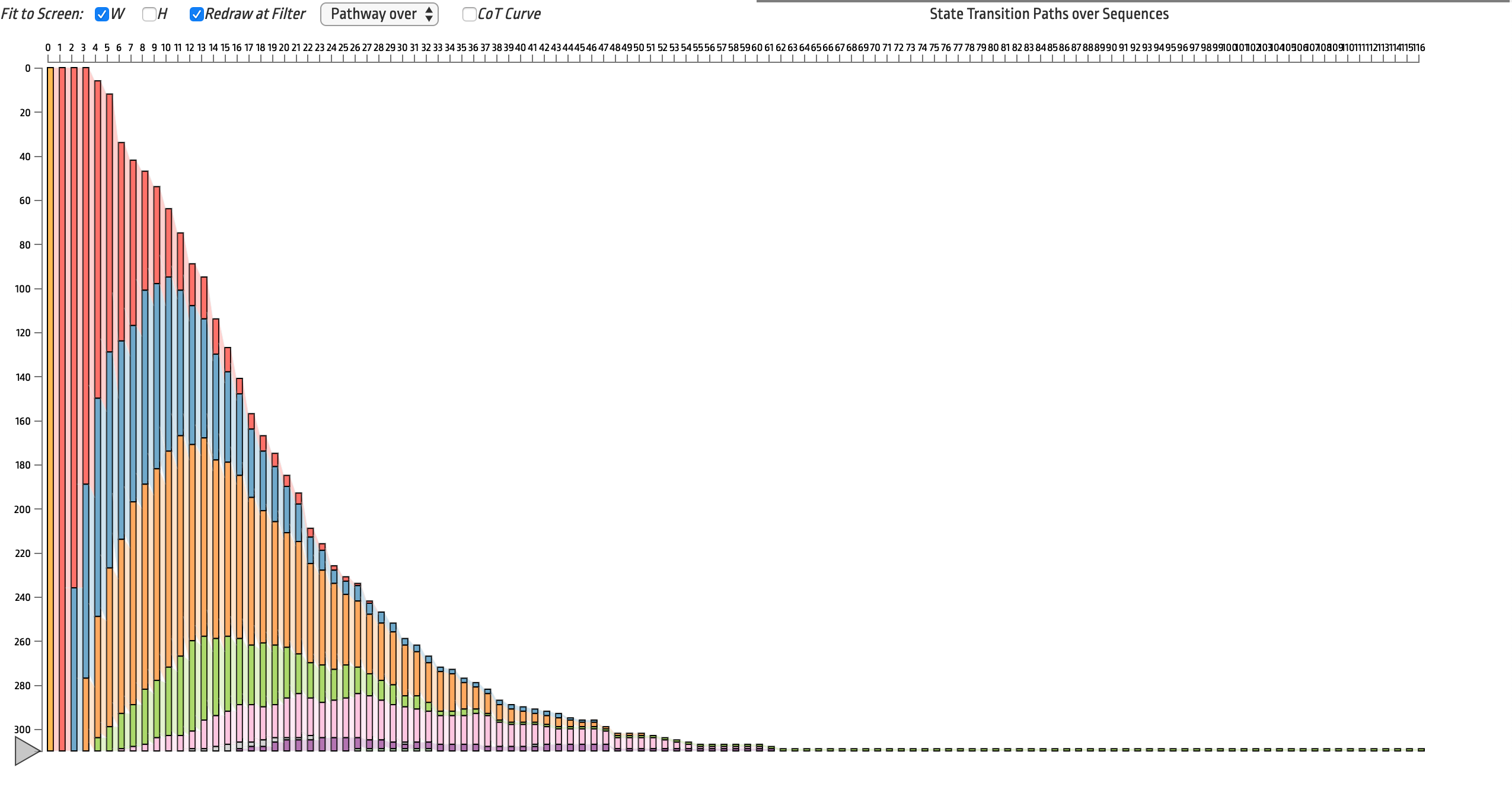}
        \vspace{-.5cm}
        \label{fig:obspathway2}
    \end{subfigure}
    \begin{subfigure}{.33\textwidth}
        \centering
        \includegraphics[width=\textwidth, height=.56\textwidth, trim={.0cm 5.25cm 8.0cm 1.2cm},clip]{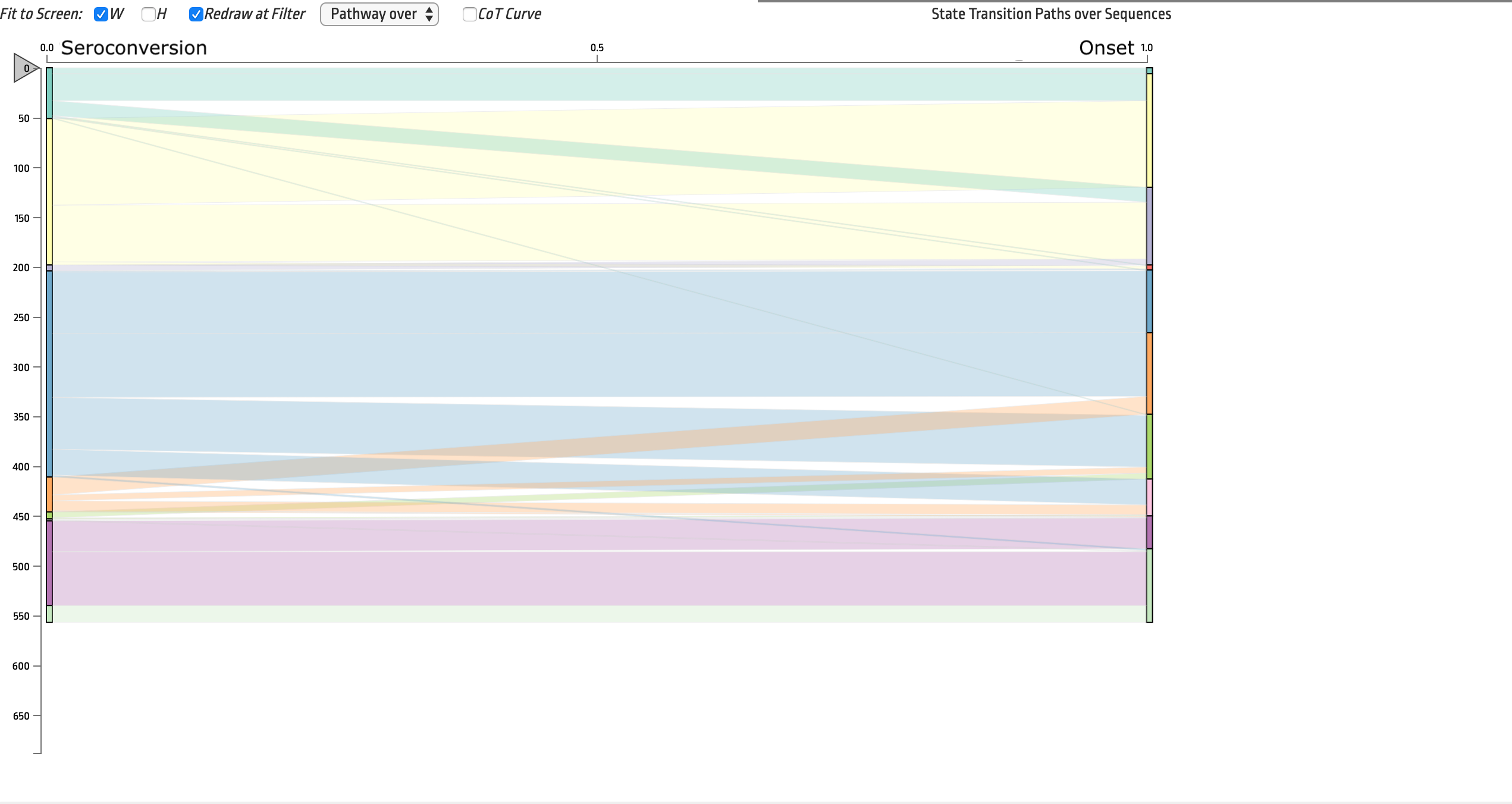}
        \vspace{-.5cm}
        \label{fig:obspathway3}
    \end{subfigure}
    \vspace{-.25cm}
    \caption{\obspathway summarizes state transition pathways for subjects over observed visits. \textbf{Left}: \alignment rearranges nodes and paths with respect to a new anchor point (between states 2 and 3), showing the proportion of State 0 decrease over first 10 visits; \textbf{Middle}: \obspathway shows pathways for subjects who started their visit with State 0 (green); the first node is highlighted in orange after users click to filter; \textbf{Right}: Users converted the view to only show start and end nodes and the path between the two nodes, showing in which state subjects seroconverted and get diagnosed (onset).}
    \label{fig:obspathway}
\end{figure*}

\subsection{\transit}

Pathway views show various representations of state transition patterns over sequences of visits or over time (\textbf{T2}).
Together with \dist, \transit (Fig.~\ref{fig:teaser}~(C)) includes views that help users to understand subjects' state transition patterns.
State transition patterns can be represented in many different ways as shown in previous studies in Section~\ref{sec:related}. \transit includes five views that users can choose from. Each view has advantages and disadvantages. In the following sections, we discuss the design of views and their trade-offs.

\subsubsection{\pathwayfall}

\pathwayfall shows subjects' state transition patterns using parallel beeswarm visualizations overlaid with individual subjects' trajectory lines.
As Fig.~\ref{fig:pathwayfall} shows, we first divide the vertical space by the number of states, each of which represents a state.
A dot indicates a visit of a subject, which is colored corresponding to its state and placed on the timepoint (X) and the state space (Y). 
In this process, we show density of visits while preserving their timestamps using the beeswarm algorithm~\cite{eklund_beeswarm_2017}.
Then, we visualize trajectory lines for individual subjects that follow the state transition pathways over time.
To reduce the clutter while preserving the transition points, we use a modified version of force-edge bundling technique~\cite{holten_force_2009}. 
To contrast the patterns more vividly, we used a dark background color.
A participant commented on the design of \pathwayfall, ``it looks like Mardi Gras beads hanging on the wall. Easy to see where (in which state) they hook and tangle up (congested departing/landing time periods per state).''

The overview in Fig.~\ref{fig:pathwayfall} (Left) shows three distinctive trajectory groups, showing state transitions: i) 0--2; ii) 3--7; iii) 8--10.
Together with state characteristics found using \dist, we observe that each of trajectory group starts with no autoantibodies.
The three trajectories show heterogeneous progression pathways with respect to their first autoantibodies: i) multi-AAB First (0--2); ii) IAA First (3--7); iii) GADA First (8--10).
We may want to create a subgroup among the IAA First (3--7) trajectory, who gains IAA early before the age of 20 months.
To achieve the goal, we apply filters by specifying transition points, as shown in Fig.~\ref{fig:pathwayfall} (Right).
The updated figure shows a onward progression pattern of the subgroups.
Those who early progressed into IAA (4) seem to further possess more autoantibodies in early ages, as shown in thicker vertical paths drawn in the leftmost area.
Users can also observe outliers who progressed beyond state 7 into states 8 and 9.

\subsubsection{\obspathway}

\obspathway allows users to gain an overview of transition patterns by comparing the height of stacks over visits first (\textbf{T2}).
\obspathway shows state transition patterns of multiple subjects using the sankey diagram approach.
As Fig.~\ref{fig:obspathway} shows, the view is made of nodes and paths.
Nodes are vertically stacked for each visit (x-axis), where each stack's height represents the number of subjects with the corresponding state at the visit. 
Paths are drawn horizontally between two consecutive visits, where each path connects a state (State-A) of the previous visit (Visit-0) to another state (State-B) of the current visit (Visit-1).
The thickness of each path represents the number of subjects who transitioned from State-A to State-B over the visits between 0 and 1.

The main advantage of this view is summary of state transitions over a series of observations. Sankey diagrams allow users to compare the volume of transitions between consecutive visits. Alignment allows users to understand the state transition patterns with respect to specific states. Fig.~\ref{fig:obspathway}~(Left) shows, by shifting the alignment to focus on the first trajectory group (Multi-AAB First), we observe that subjects in the trajectory tend to have zero autoantibody (green; state 0) in early age, many of them gain multiple autoantibodies (yellow; state 1) as they grow, but rarely lose GADA and IAA (purple). Furthermore, users can simplify the state transitions using the biparite Sankey diagram so that they can focus on the relationship between state transitions and health outcomes. Fig.~\ref{fig:obspathway}~(Right) shows sankey diagrams that summarize in which states subjects show seroconversion (left) and onset of type 1 diabetes (right). The first autoantibody state per each trajectory, namely state 1 (yellow), state 4 (blue), and state 9 (purple), are the most common states for seroconversion. On the other hand, subjects get diagnosed with type 1 diabetes in various states. The downside of this approach is that it aggregates individuals, so users cannot view or choose individual patients from this view.

\subsubsection{\timepathway}

\begin{figure}[t!]
    \captionsetup[subfigure]{labelformat=empty}
    \centering
    \begin{subfigure}{.24\textwidth}
        \centering
        \includegraphics[width=\textwidth, frame, trim={.0cm .5cm 12.0cm 1.2cm},clip]{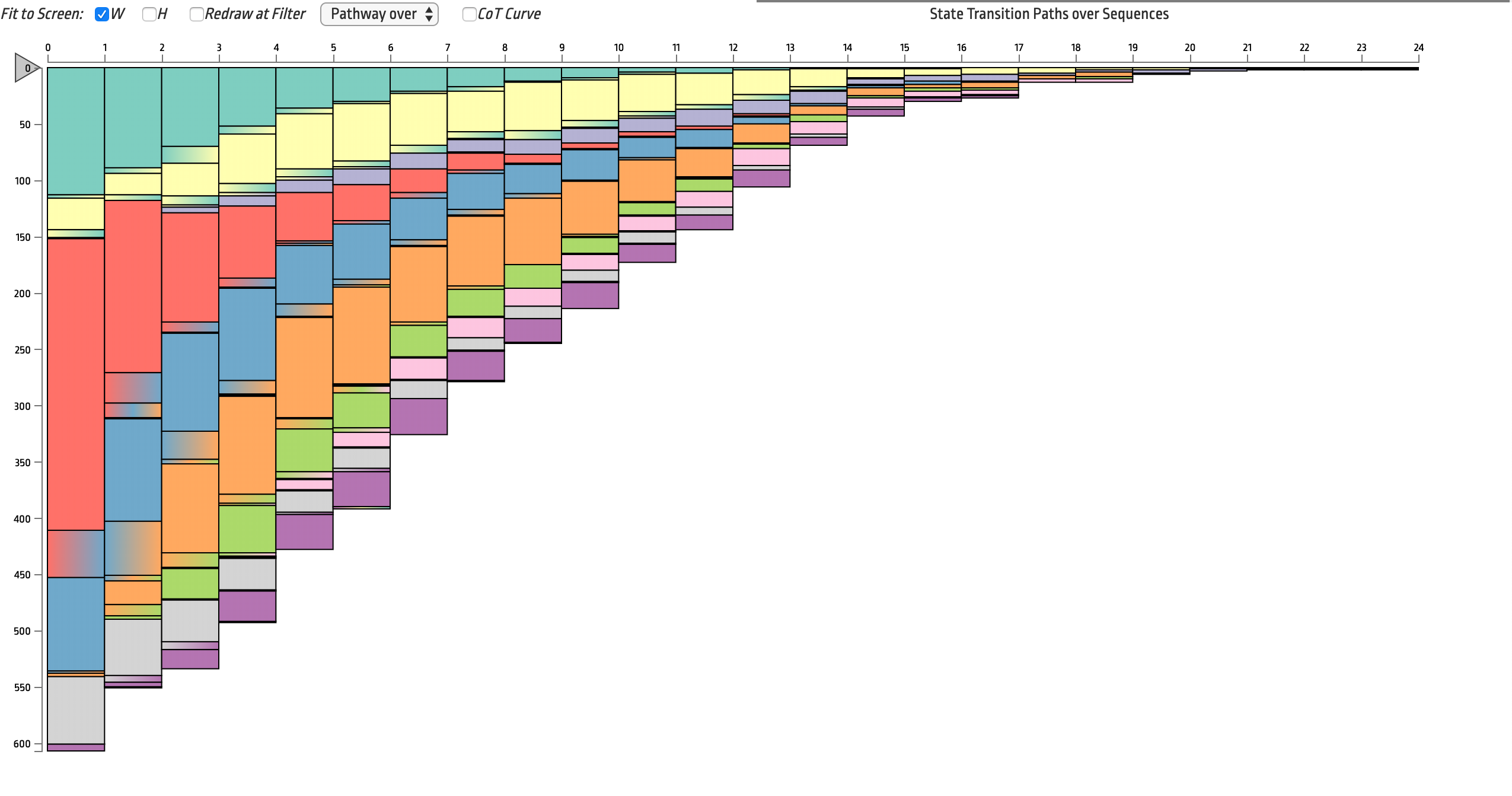}
        \vspace{-.5cm}
        \label{fig:timepathway1}
    \end{subfigure}
    \begin{subfigure}{.24\textwidth}
        \centering
        \includegraphics[width=\textwidth, frame, trim={.0cm .5cm 12.0cm 1.2cm},clip]{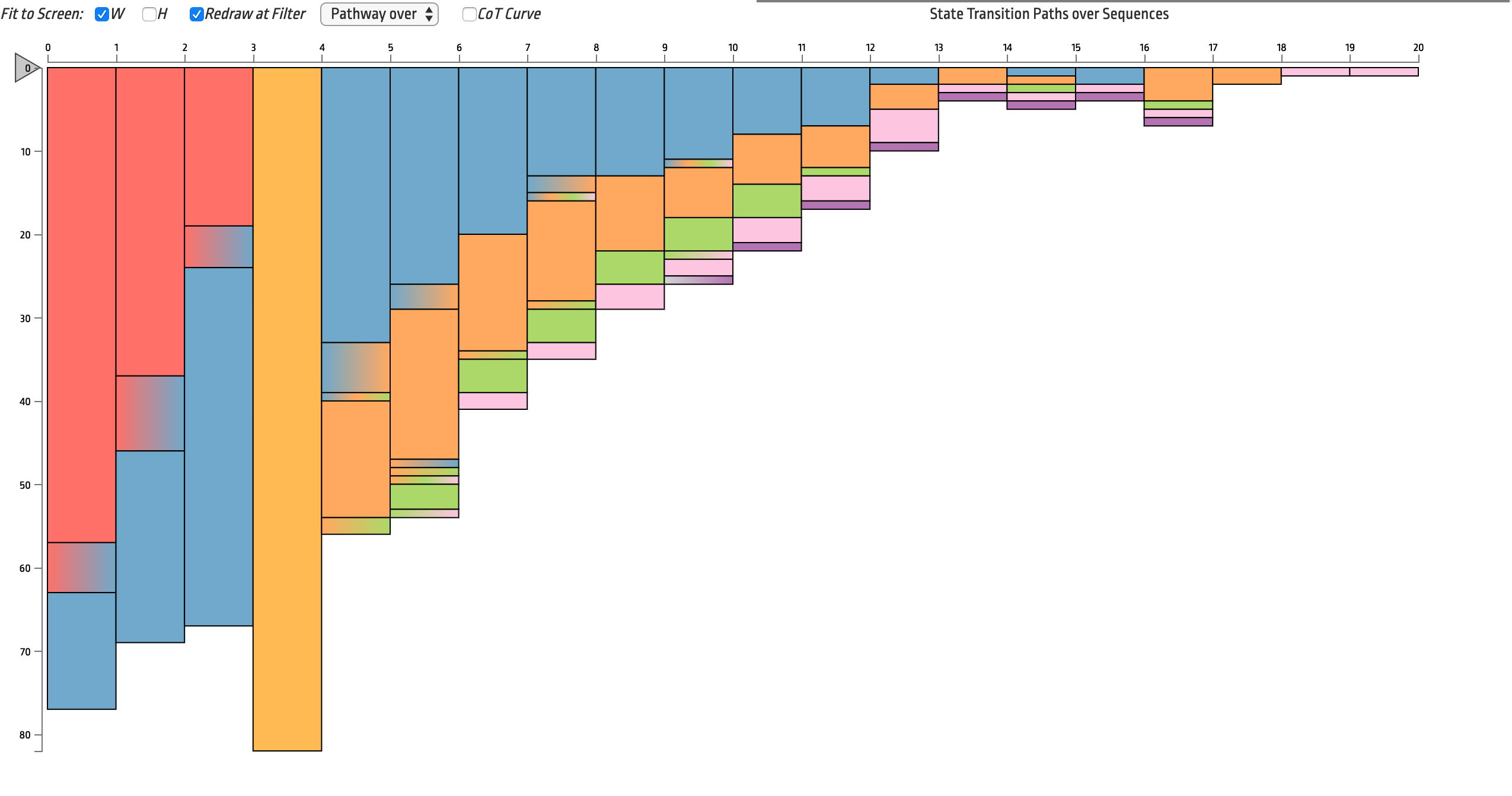}
        \vspace{-.5cm}
        \label{fig:timepathway2}
    \end{subfigure}
    \caption{\timepathway shows state transition pathways. \textbf{Left}: \timepathway shows stacked bars aligned at an anchor point between states 3 and 4; \textbf{Right}: Users set a filter by clicking on State 4 at subjects' age between 4 and 5 years old.}
    \vspace{-.5cm}
    \label{fig:timepathway}
\end{figure}

\timepathway aggregates state transitions with respect to a time unit of users' choice.
Thus, the patterns that one can observe are similar to \obspathway, but it shows the temporal patterns of state transitions.
In Fig.~\ref{fig:timepathway} (Left), the overview shows state transition patterns aggregated per subjects' age in `years'.
State 3 (red), which represents no autoantibody, is the most prevalent state for subjects.
It is interesting to observe that some subjects show early transition from no autoantibody (State 3) to IAA (State 4) before one year old, shown as a stack with red-to-blue gradient on the leftmost bar.
We can set filters by showing subjects who went through State 4 (blue) at their age between 4 and 5 years old as Fig.~\ref{fig:timepathway} (Right) shows.
The majority of subjects possessed IAA at State 4 a year before and many of those did not lose or gain a new autoantibody afterward, as seen from bars in blue throughout the observed periods.
The advantages and disadvantages of this approach are similar to those of \obspathway. One distinguished advantage is to be able to show state transition with respect to ages rather than visits, which is often more relevant to clinical observational research.

\subsubsection{\chord}

\begin{figure}[b]
    \centering
     \vspace{-.5cm}
    \begin{subfigure}{.24\textwidth}
        \centering
        \includegraphics[width=.5\textwidth, trim={.0cm .0cm .0cm 0.0cm},clip]{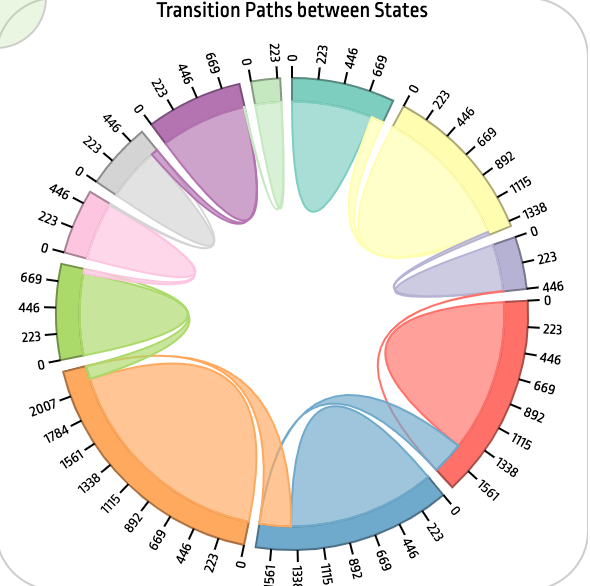}
        \label{fig:chord1}
    \end{subfigure}
    \begin{subfigure}{.24\textwidth}
        \centering
        \includegraphics[width=.5\textwidth, trim={.0cm 0.0cm .0cm 0.0cm},clip]{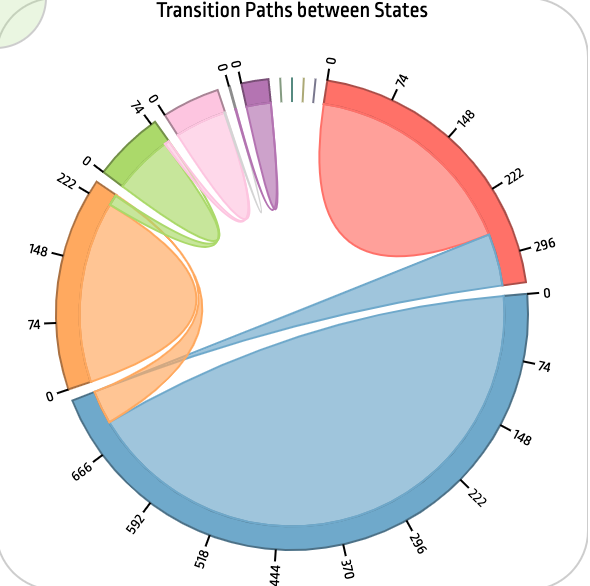}
        \label{fig:chord2}
    \end{subfigure}
    \vspace{-.25cm}
    \caption{\chord shows a summary of state transitions between all pairs of consecutive states.\textbf{Left}: It summarizes state transitions for all subjects; \textbf{Right}: It shows subjects who had State 4 between 3 and 4 years old.}
    \label{fig:chord}
\end{figure}

\chord shows the overall state-to-state transition patterns over all consecutive pairs of observations from subjects using a radial network diagram.
The view can be used to understand the overall frequency of state transitions (\textbf{T2}).
This view serves as an explanatory summary of the state transition learned from the particular setting of HMMs.
It allows users to compare the number of visits per state by viewing the size of the nodes (states) on the circular layout and to compare the transitions between two states by the size of arcs connecting two nodes inside the circular layout.
% In case of forward-only progression models, one arc is drawn per pair of different states.
As Figure~\ref{fig:chord} (Left) shows, state transitions occur rarely.
On the other hand, a cohort with subjects who had IAA with State 4 (blue) between ages of 3 and 4, as shown in Fig.~\ref{fig:chord} (Right) shows a greater proportion of transitions among States 3 (red), 4 (blue), and 5 (orange) than the rest of states.
Users can filter subjects who contain specific subpatterns by clicking an edge of interest \textbf{T5}.
\chord summarizes state transitions by frequency. The summary is useful for users to determine the overall prevalence of states and their transitions. Such patterns cannot be easily extracted in other views. On the other hand, the view does not provide any temporal context (age) of state transitions. Thus, the view is useful along with other views like \pathwayfall.

\subsubsection{\freq}

\begin{figure}[bt]
    \centering
    \begin{subfigure}{.235\textwidth}
        \centering
        \includegraphics[width=\textwidth, frame, trim={.0cm 2.3cm 6.0cm 1.2cm},clip]{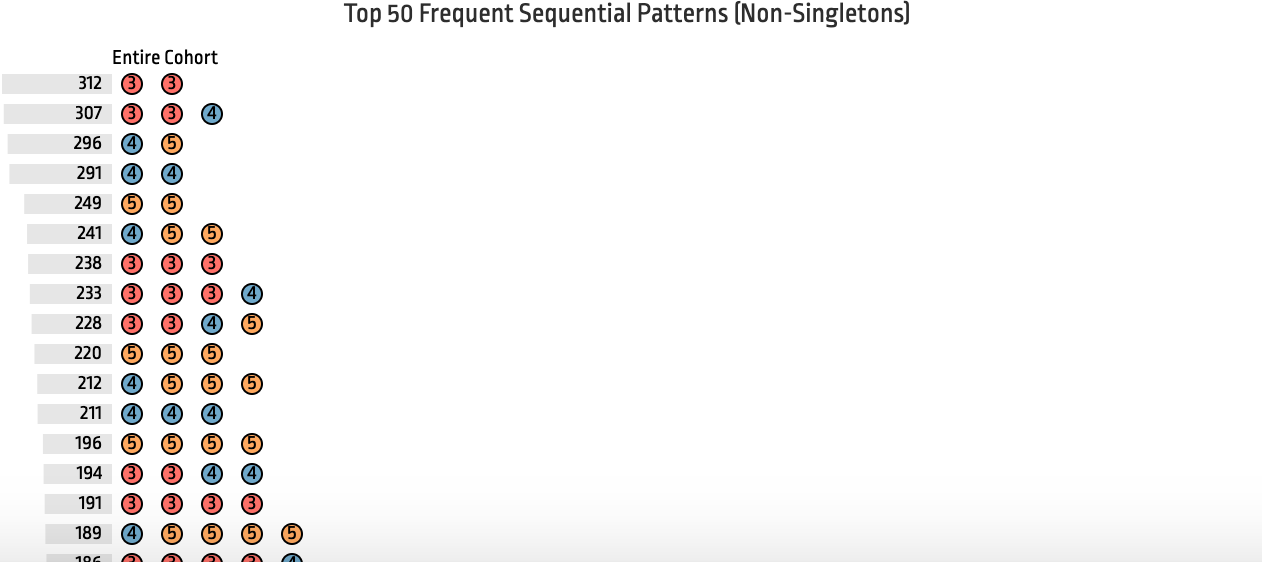}
        \label{fig:freq1}
    \end{subfigure}
    \begin{subfigure}{.235\textwidth}
        \centering
        \includegraphics[width=\textwidth, frame, trim={.0cm 2.3cm 6.0cm 1.2cm},clip]{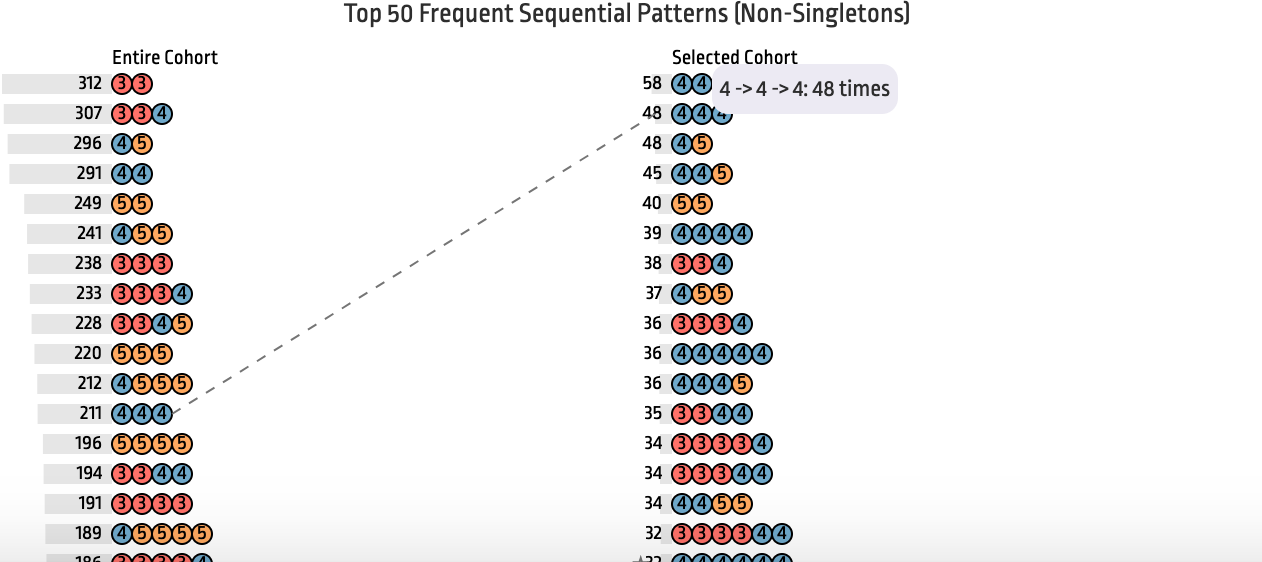}
        \label{fig:freq2}
    \end{subfigure}
    \vspace{-.25cm}
    \caption{\freq shows the result of pattern mining algorithm (BIDE)~\cite{wang_bide_2004}. \textbf{Left}: The top-50 most occurring state transition patterns; \textbf{Right}: A new pattern list for the cohort who were breastfed.}
    \label{fig:freq}
    \vspace{-.65cm}
\end{figure}

\freq in Fig.~\ref{fig:freq} shows a list of frequently occurring state transition patterns mined from all consecutive observations (\textbf{T2}).
By running the BIDE algorithm~\cite{wang_bide_2004} on subjects' state sequences, we first generate the most frequently occurring state sequence patterns.
To avoid filling up the list with fine granular patterns, we filtered top 50 patterns in terms of the number of unique subjects with the corresponding pattern; we also do not take into account any single-state patterns.
Then, we list all patterns, where each pattern is a row of state circles, in a vertical order, as shown in Fig.~\ref{fig:freq} (Left).
Inspired by Peekquence~\cite{kwon_peekquence_2016}, we also visualized the frequency of the patterns as a bar chart next to each pattern.
As Fig.~\ref{fig:freq} (Right) shows, once users set filters for those who possessed IAA (state 4 in blue) between 4 and 5 years old, the view expands and introduces yet another list of patterns.
By doing so, users can check the ranked list of patterns for those who possess and maintain IAA at early age.

As discussed, each view provides a unique perspective of state transition patterns. It is impossible to show the different perspectives in a single view, which motivates the options we have provided. Furthermore, different views enable opportunities for refining  subgroups. For instance, users can create a subgroup by setting filters on \pathwayfall, \chord, and \freq. In the design process, we learned that supporting subgroup building activity is facilitated by multiple, different views, which we will revisit in Section~\ref{sec:discussion}.

\subsection{\subjecttimeline}

\subjecttimeline (Fig.~\ref{fig:teaser}~(E)) shows individual subjects' observations over time (\textbf{T3}). The panel includes two views: 1) \kernel; 2) \subjectlist.

\subsubsection{\kernel}

\begin{figure}[tb]
    \centering
    \includegraphics[width=.485\textwidth, frame, trim={.0cm 1.0cm .0cm 0.15cm},clip]{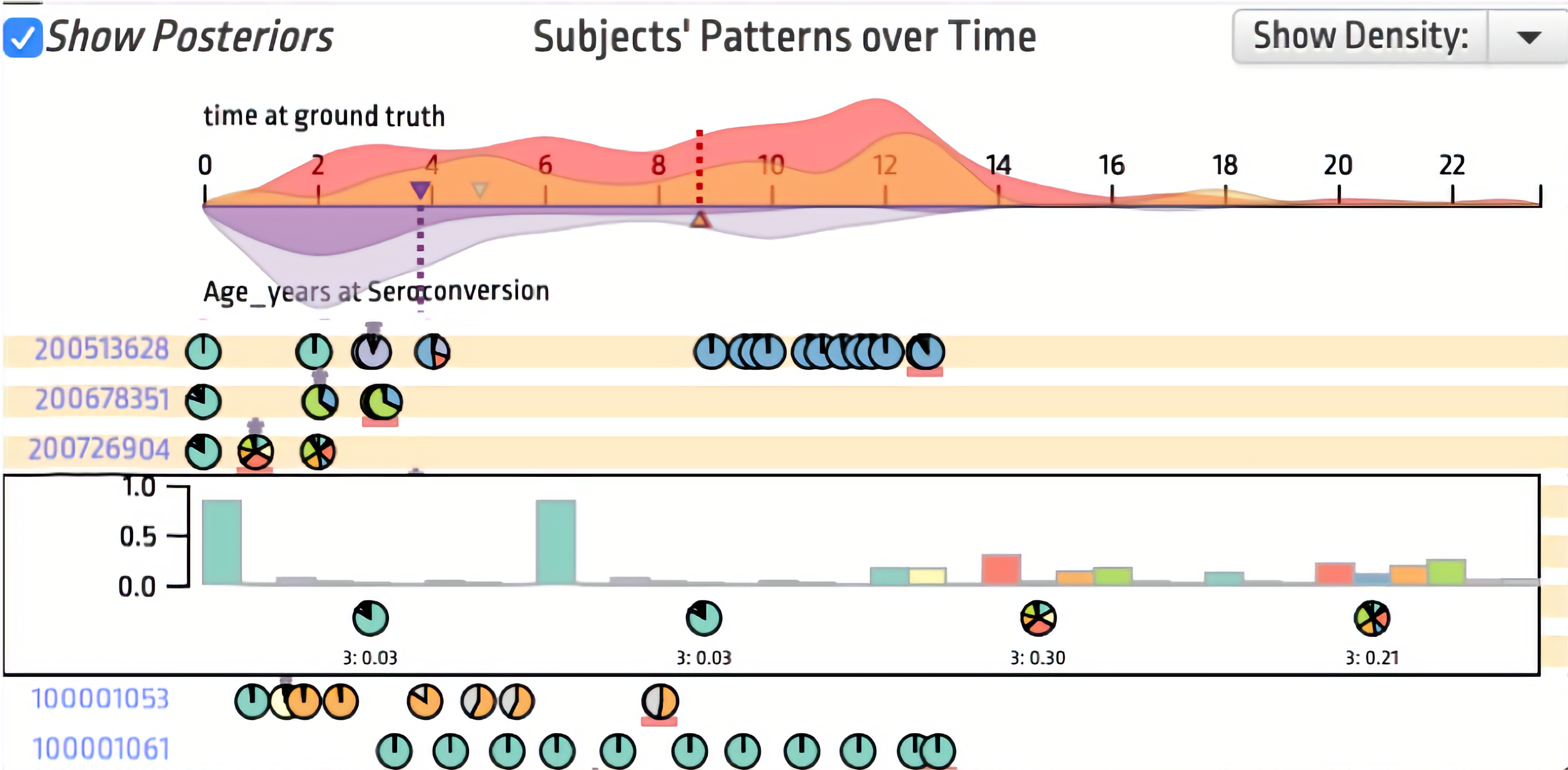}
    \caption{\subjecttimeline shows 1) \kernel and 2) \subjectlist. \kernel overlays additional dual Kernel Density Diagrams for a selected cohort, and \subjectlist highlights subjects in the cohort. Users can open a popup to view posterior distributions.}
    \label{fig:subjecttimeline}
    \vspace{-0.5cm}
\end{figure}

\kernel shows the two independent 1-D Kernel Density (KD) diagrams, each with increasing y-axis (density) in the opposite direction, upward and downward, respectively, as Fig.~\ref{fig:subjecttimeline} shows.
The KD above the timeline shows the density of time at the onset (e.g., diagnosis of type 1 diabetes) (\textbf{T3}).
Subsequently, the KD below the timeline can be freely chosen by users, with the dropdown menu on the top-right corner, among the secondary onset variables (e.g., seroconversion for type 1 diabetes) (\textbf{T3}).
Each KD includes an arrow indicating the mean age of its onset for subjects.
Once users set filters, the KDs introduce and overlay brighter KDs inside, which indicate KDs for the selected cohort (\textbf{T5}).
Fig.~\ref{fig:subjecttimeline} shows the cohort who were never brestfed.
The density chart in orange above the timeline suggests that the cohort' mean diagnosis age is at 8-9 years old with a peak around age 12 and a long tail towards older ages.
On the other hand, the purple density below the timeline indicates that the cohort has earlier seroconversion than the overall population (light purple)  (\textbf{T3}).

\subsubsection{\subjectlist}

By viewing the state sequences of individual subjects in \subjectlist, users can inspect the details at the patient level (\textbf{T3}).
\subjectlist provides a list of subjects' observations as dots (visits), color-coded based on the state, over a timeline.
For each subject, we draw a narrow, red rectangle below a dot for a visit with the onset of the disease and an asterisk symbol above a visit with the onset of the secondary onset variable (\textbf{T3}).
In \subjectlist, users can choose to show the posterior distribution over all states as a pie chart.
By observing how slices of pies change their values over time, users can learn states of visits, inferred by the HMM (\textbf{T1}).
Fig.~\ref{fig:subjecttimeline} shows the cohort of patients who were never breastfed.
Most dots present full circles (dominant with a slice), which show model's confidence in the state assignments.
However, a subject on the popup shows evenly-spread posterior distrbutions over 11 states for the last two visits, which signify uncertainties of state assignment for those visits. 
The subject shows seroconversion (purple star above pie) and diagnosis (red underline below pie) at the same visit before 1 year old.
Both onsets rarely occur at the same visit, which is why the model thinks of the visits as `uncertain' with posterior values spread over multiple states.

\subsection{\static}

\static (Fig.~\ref{fig:teaser}~(A)) shows a list of selected measures that do not change over time in horizontal bar charts.
Users can use this view to show the general summary of subjects in terms of their static measures, such as genetic profile, gender, race, and family history status.
Using this view, users can build cohorts based on factors related to subjects' genetic characteristics and biological attributes (\textbf{T4}).
For example, users can click on 'Y' on the 'bfever' bar chart and 'F' on the 'SEX' bar chart to create a cohort that includes only female subjects who were breastfed.

\subsection{\subgroup: Build, Refine, Compare, Retain}

In this view group, users build, refine, and maintain subject groups using a variety of operations.

\subsubsection{\cohort}

\cohort (Fig.~\ref{fig:teaser}~(G)) shows a list of subject cohorts that have been built and kept by the user.
As mentioned in previous subsections, users can build cohorts using any of the views described above; whenever users make updates on the cohorts, the changes are automatically saved in this view (\textbf{T5}).
The height of each bar corresponds the relative number of subjects within each cohort.
The title can be modified with free text so that users can keep short descriptions about the cohort.
As users hover over each bar, a popup view shows the description of filters, namely a list of pairs of an attribute name and its value range, used to create the corresponding cohort.
Users can also export and import cohorts from existing static variables by choosing one variable on the dropdown menu at the top-right corner (\textbf{T5}).

\subsubsection{\query}

\begin{figure}[t]
    \centering
        \includegraphics[width=.5\textwidth, frame, trim={.2cm 0.2cm .2cm 0.0cm},clip]{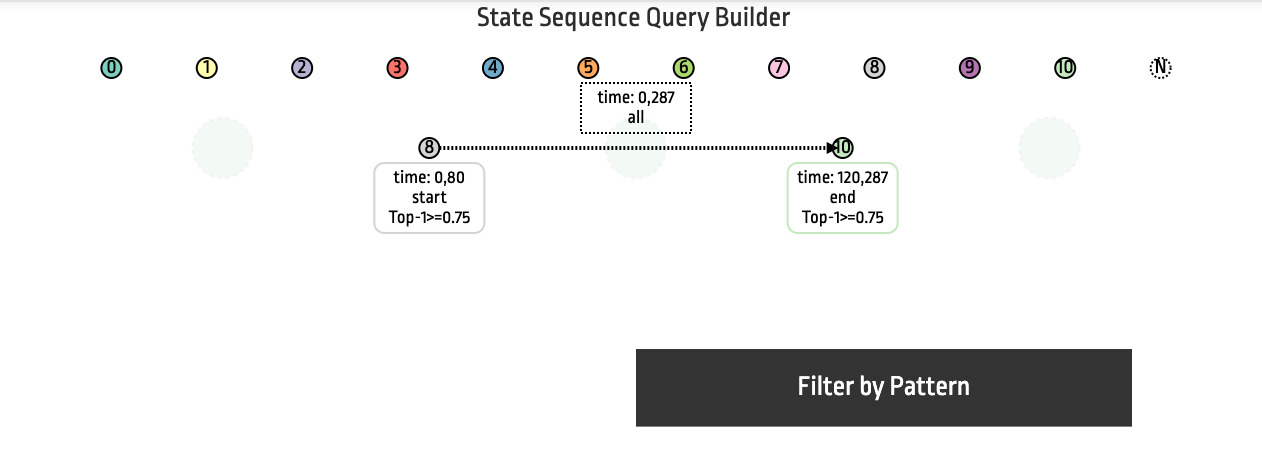}
    \caption{\query allows users to build and refine state transition patterns. Users can adjust filtering criteria (e.g., time of visit) for each node and edge on a context menu.}
    \label{fig:query}
    \vspace{-.5cm}
\end{figure}

\query is a way users can freely define and refine queries to create a cohort based on state transition patterns (\textbf{T5}).
We built this view based on users' feedback. Users want to build cohorts by setting sophisticated filters using state transition patterns. We initially considered a command-line interface, where users can `type' to build a query using some syntax language like SQL. 
We later found that our users can face difficulties in learning how to use the syntax language.
Thus, we decided to build a graphic user interface that allows users to interactively build state sequence query.
As Fig.~\ref{fig:query} shows, the view includes a canvas, where users can graphically create transitions of interest by using state nodes.
The view initially shows an empty canvas with horizontally arranged state nodes with an additional dummy node (in total, $N$+1 nodes).
Using a drag-and-drop interaction, users can express the state transition patterns that they are looking for.
Though the node-link representation is intuitive to understand and learn how to build, it may sacrifice expressivity.
Users may want to refine their search based on other attributes related to states (nodes) or transitions (edges). 
For example, instead of just state 4, users may want to filter state 4, which occurred in specific ages.
Thus, we added a popup menu for users to express more sophisticated constraints on states and transitions.
Every node and edge has its own properties shown with labels, which can also be edited on a popup window.
Each node has three properties: 1) time, 2) node\_at, and 3) posteriors, and each edge between two nodes has two properties: 1) the maximum time between nodes and 2) the sequential order of the two nodes.
Once users configure patterns, they hit the ``Filter by Pattern'' to execute the query.
Fig.~\ref{fig:query} shows that a user built a query using two nodes: state 8 (no autoantibody) and state 10 (GADA and IA2).
The pattern includes the following configurations: 1) State 8 occurs between age 0 and 80 months at the beginning visit of the subject; 2) State 10 occurs between age after 120 months at the final visit; 3) both states are assigned with posterior probabilities greater than or equal to 0.75.

\begin{figure*}[t]
   \centering
    \begin{subfigure}{.49\textwidth}
        \centering
        \includegraphics[width=\textwidth, trim={.0cm 0.0cm 0.0cm 0cm},clip]{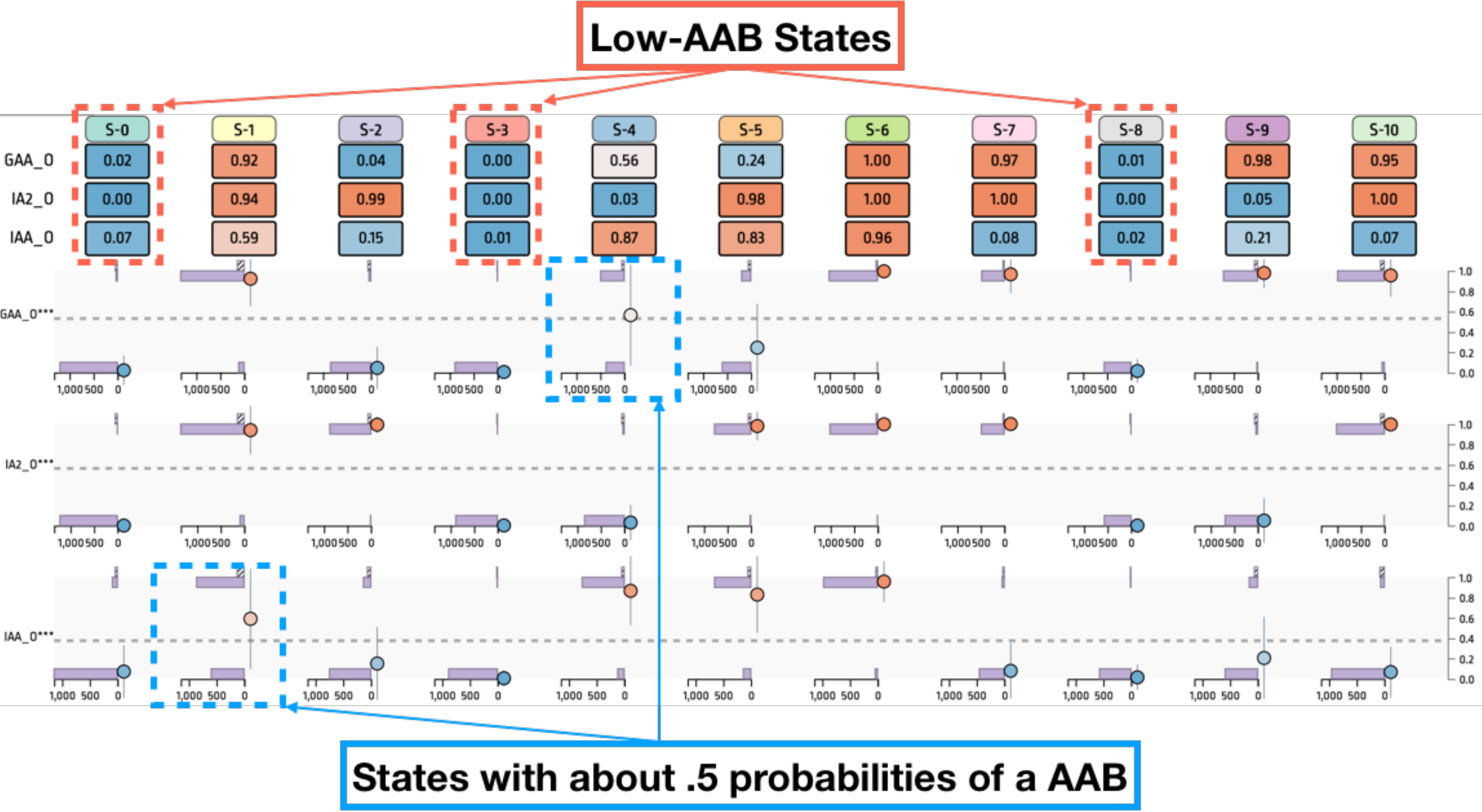}
        \caption{State characteristics for 11-state model are summarized.}
        \label{fig:casestudy0}
    \end{subfigure}
    \begin{subfigure}{.49\textwidth}
        \centering
        \includegraphics[width=\textwidth, trim={.0cm 0.0cm 0.0cm 0cm},clip]{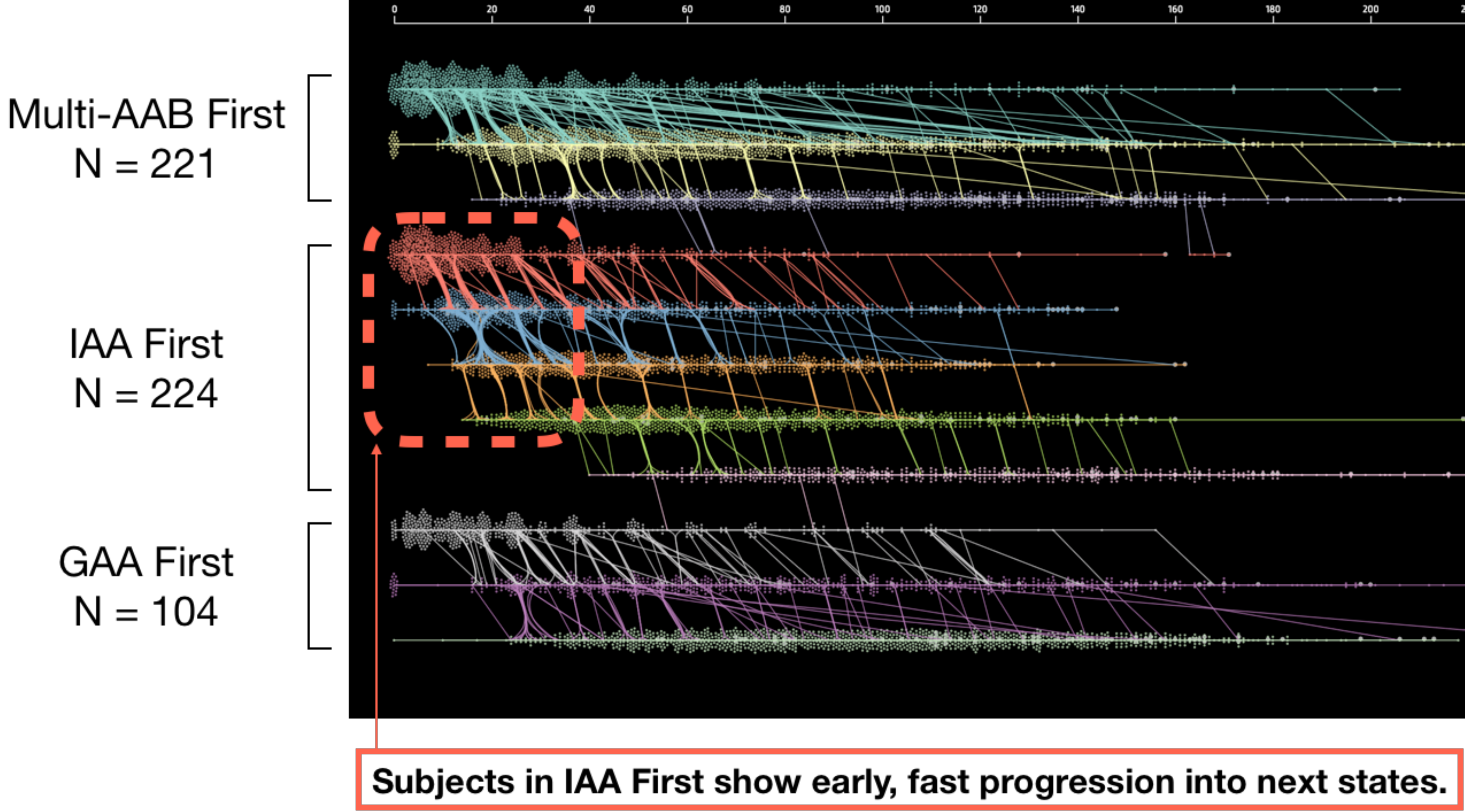}
        \caption{Three distinct trajectory subgroups are revealed.}
        \label{fig:casestudy1}
    \end{subfigure}
    \begin{subfigure}{.49\textwidth}
        \centering
        \includegraphics[width=\textwidth,height=.6\textwidth, trim={.0cm 0.0cm 0.0cm 0cm},clip]{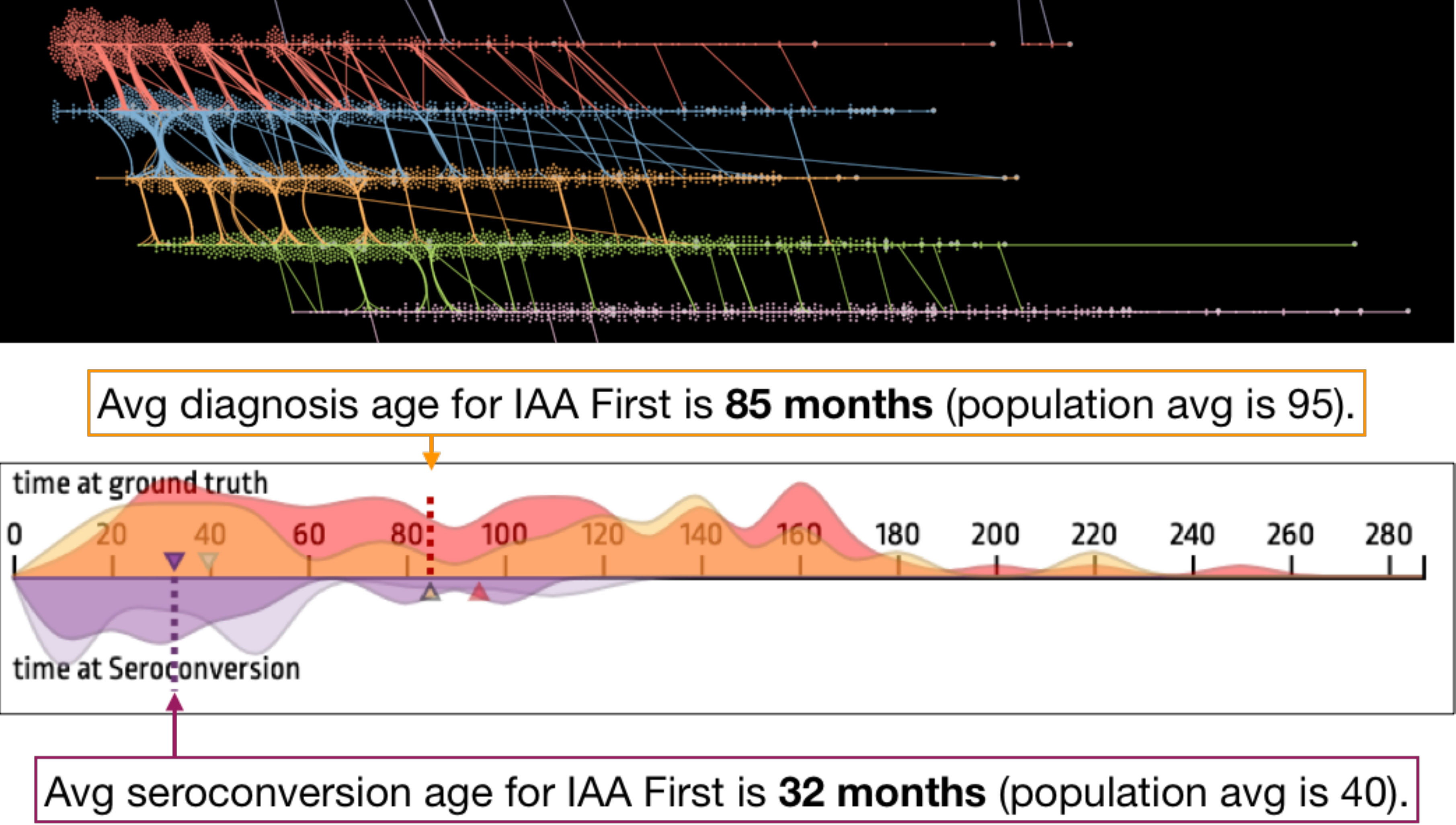}
        \caption{Average ages for diagnosis and seroconversion are shown.}
        \label{fig:casestudy2}
    \end{subfigure}
    \begin{subfigure}{.49\textwidth}
        \centering
        \includegraphics[width=\textwidth,height=.6\textwidth, trim={.0cm 0.0cm 0.0cm 0cm},clip]{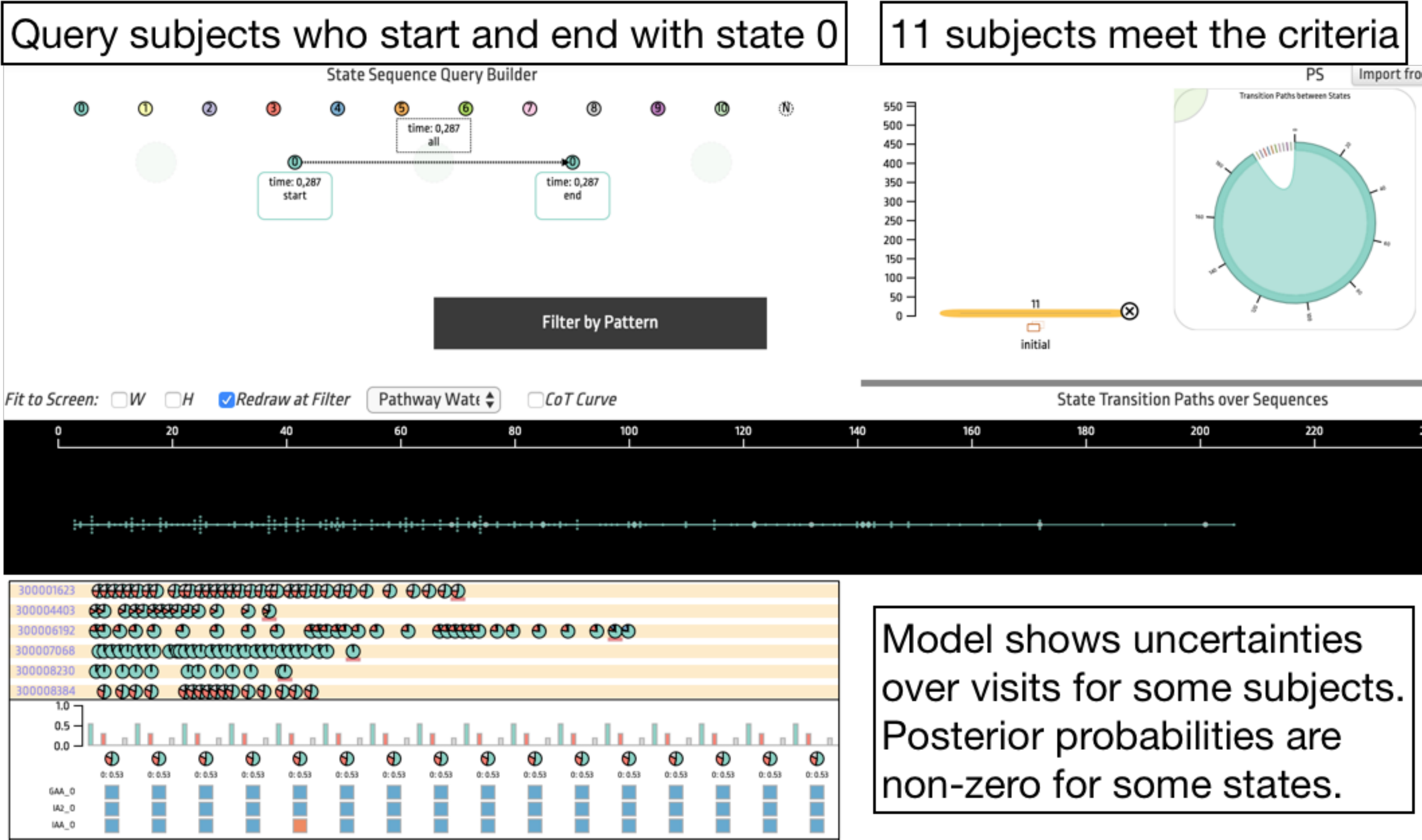}
        \caption{11 outliers are queried and investigated in detail.}
        \label{fig:casestudy3}
    \end{subfigure}
    \caption{The figure shows how users explore disease progression patterns using \toolname following the scenario in Section~\ref{sec:usage_scenario}.(a) Users view and compare characteristics of states (\textbf{T1}); (b) users gain an overview of state transition patterns (\textbf{T2}); (c) users probe the relationship between health outcomes and disease progression trajectories (\textbf{T3}); (d) users check details of subjects and manage subgroups (\textbf{T4, T5}).}
    \label{fig:casestudy}
    \vspace{-.5cm}
\end{figure*}

\vspace{-.25cm}
\section{Clinical Studies of Type 1 Diabetes}
\label{sec:casestudy}
In this section, we introduce 1) a usage scenario that demonstrates how \toolname help users explore disease progression pathways and 2) users' experiences of using \toolname.

\vspace{-.25cm}
\subsection{Long-Term Collaboration with Study Group}
\label{sec:collab_study_group}

This section briefly explains the long-term collaboration with clinical researchers in type 1 diabetes, who contributed to the user experiences and usage scenarios.
As part of the \textit{T1DI} study group, we organized four quarterly workshops in four different cities in the US and Europe between October, 2018 and October, 2019.
The goal of each workshop was to gain clinically meaningful insights into subjects' trajectories prior to diagnosis.
In each workshop, clinical researchers and two leading authors of the paper used \toolname T1D to explore T1D data collected from multi-site birth cohort studies.
During the workshop, we conducted Pair Analytics~\cite{arias2011pair}, where the two leading authors drive the tool and clinical researchers interpret the patterns, formulate hypotheses, and derive clinically meaningful insights.
As a result, we derive the usage scenario that leads to key insights, which will be presented to an upcoming Diabetes conference, in Section~\ref{sec:usage_scenario}.
We recruited nine clinical researchers, who actively participated in the workshop sessions, for the interview session to ask questions about their experiences.
In the open-ended discussion with each, we focus our discussion on the cost and value of DPVis for clinical research.
We recorded the conversation, which took an hour on average, and coded the transcribed scripts using grounded theory~\cite{muller_curiosity_2014}.
We report the results in Section~\ref{sec:user_feedback}.
Though the study is part of large study consortium, the process and outcome reflect the goal of Multi-dimensional In-depth Long-term Case study (MILC)~\cite{shneiderman_strategies_2006}.

\vspace{-.25cm}
\subsection{Usage Scenario}
\label{sec:usage_scenario}

Our users investigated disease progression patterns from 559 subjects who were diagnosed with type 1 diabetes (T1D), using observational data from three prospective clinical studies: DiPiS~\cite{jonsdottir_childhood_2018}, DAISY~\cite{rewers_newborn_1996}, and DIPP~\cite{nejentsev_population-based_1999}.
Researchers decided to model the subjects' progression data using Continuous time HMM (CT-HMM), a variant of HMM.
They trained models with a range of parameters, primarily the number of states, from 2 to 20.
They focused on autoantibodies (AAB) variable due to recent interest and work in the type and number of AAB development in the T1D community. 
For training the autoantibody (AAB) model, they used three observation variables, all markers of disease progression: glutamate decarboxylase autoantibody (GADA), insulin autoantibody (IAA), and islet autoantigen-2 autoantibody (IA2A) (More studies on autoantibodies can be found in~\cite{torn_glutamic_2000}).
To test their hypothesis, they trained the models with a range of states ($N$ $=$ [2, 20]).
After testing the models with multiple rounds of cross-validations, they narrowed down to the best performing model based on the log-likelihood scores on the validation dataset: an 11-state autoantibody (AAB) model.

What are the states discovered by the 11-state model (\textbf{T1})?
Users open \heatmap and \dist to interpret the states with respect to the evolution of AABs.
They found that there are at least three low-AAB states, which are states 0, 3, and 8, highlighted in red in Fig.~\ref{fig:casestudy0}. 
They are characterized as buttons in blue for the majority of rows in \heatmap, including the three AABs, which indicate that the three states represent young, early states prior to onsets of any AAB.
The remaining states can be categorized into groups with respect to the number of AAB: (i) 1-AAB state: 2, 4, and 9; (ii) 2-AAB state: 4, 7, and 10; (iii) 3-AAB state: 1 and 6.
Users found that state 4 can be categorized into 1-AAB, 2-AAB, or both because it has .56 probability in GADA; state 2 can be categorized into 2-AAB, 3-AAB, or both because it has .59 probability in IAA, as Fig.~\ref{fig:casestudy0} highlights in blue.
This indicates that transitions including state 2 or 4 can also be subdivided into more granular trajectory groups.
After inspecting the state characteristics, users confirmed that the 11-state model expose diverse states represented by various combinations of AABs.

Together with the characteristics of individual states, users set out to understand disease progression patterns (\textbf{T2}) and their relationship with onsets of diagnosis (\textbf{T3}).
The overview shows that there are three large distinctive trajectory groups, each of which follow state transitions: i) 0-2, ii) 3-7, and iii) 8-10, which are named as ``Multiple-AABs First'', ``IAA First'', and ``GAA First'' groups, respectively, as Fig.~\ref{fig:casestudy1} shows.
The names are given as above by clinicians because the very first AAB(s) subjects gained provide clinically meaningful implications.
Using \query, users create subgroups in \cohort for detailed investigation (\textbf{T5}).
The number of subjects for the three trajectory groups are 221, 224, and 104.
There are 10 subjects who do not belong to one of three because they show transitions across multiple trajectory groups: 7 subjects belong to Multiple-AABs First and IAA First, and 3 subjects belong to IAA First and GAA First.
Looking at the overview of transitions, users notice key differences between three trajectory groups with respect to transition ages.
\pathwayfall shows that the vertical paths drawn in the ``IAA First'' group are very dense especially early in their ages in comparison to the other trajectory groups.
To investigate the association between average age for diagnosis and seroconversion and different trajectory groups, users view \kernel that shows density charts for diagnosis (red) and seroconversion (purple).
Fig.~\ref{fig:casestudy2} shows that the ``IAA First'' group has earlier seroconversion and diagnosis than the population.

Following the trajectories, they found that some subjects were assigned State 0 (no autoantibodies) for all of their visits, yet they were diagnosed with T1D.
To investigate this cohort further, they queried those who started and ended their observations with State 0 using \query (\textbf{T4}), as Fig.~\ref{fig:casestudy3} shows.
The query returned 11 subjects, who were diagnosed around 119 months (10 years) old, according to \kernel.
Among the 11 subjects, 6 of them seroconverted with IAA, and the rest had no seroconversion, according to \static.
\subjectlist shows posterior probabilities over 11 states for each visit, and there are many visits that show nontrivial posterior values for states other than state 0 (\textbf{T4}).
Based on the patterns, users speculated two interpretations. 
First, the negative autoantibody at the last visit may be measurement error.
Second, the subjects may represent an endotype characterized by factors that are not captured in our data~\cite{battaglia_understanding_2017}.
They were intrigued and requested more information.

\vspace{-.25cm}
\subsection{User Experiences on \toolname}
\label{sec:user_feedback}
\vspace{-.25cm}

This section reports the summary of interviews with nine clinical researchers who participated in the workshop sessions.
Eight out of nine researchers had 10 or more years of experience (min: 3 years; max: 40 years) with clinical or healthcare research.
Five out of nine researchers were practicing medical doctors and clinical researchers, who had Doctor of Medicine in various disciplines related to diabetes research; The rest had degrees in statistics, computer science, and industrial engineering and had experiences with clinical and medical informatics. 
They had some knowledge and experiences with HMMs and visualizations in their research. 

\textbf{\toolname reduces the effort required to analyze observational data by providing interactive subgroup analysis.} Participants reported that the major benefit of \toolname is to save time and effort for analyzing observational data.
P2 mentioned that understanding disease progression patterns (\emph{T1-2}) is ``definitely doable without \toolname, but not easy, very difficult. Using \toolname makes it so much easier to do so.''
P9 mentioned that it helps her ``quickly test an idea about whether my hypotheses are worth investigating.''
Participants described the utilities that \toolname uniquely provides.
First, participants appreciated that the tool allows them ``to see the entire cohort and to single out individuals at the same time'' (P4).
Second, it was regarded useful to be able to ``build cohorts without scripts'' (P1).
The cohort building activity (\emph{T5}) allows users to validate whether ``what one has published or investigated earlier makes sense in other data, graphically'' (P7).
Third, participants were able to conduct sensitivity analysis repeatedly by refining subgroups and extracting patterns from data (\emph{T5}): ``You can conduct sensitivity analysis again and again. You can slightly change what you look at. That's a real strength. [...] It lets users test the assumptions on particular subgroups. Even kinda test the robustness. Once you identify a pattern, then you could ask: it this pattern robust even after excluding a particular study [dataset]?'' (P3).

\textbf{\toolname helps users understand HMMs transparently.} Despite the perceived usefulness of HMMs, participants had concerns about how to interpret them, but \toolname was helpful to mitigate the issues and to use the model in a more transparent manner.
HMMs are generally regarded as useful to characterize disease progression patterns of individual patients (\emph{T1}).
P1 described the usefulness as: ``HMMs are useful to understand the severity of diseases. It quantifies disease states and lets users find patterns.''
However, participants were aware of limitations with regards to how to interpret the outputs.
HMM outputs alone were not perceived as intuitive because ``there are always uncertainties in state assignment'' (P7) and ``states seem like black boxes'' (P8).
It is also challenging to interpret model outcomes without considering and keeping in mind how the data were collected: ``Censoring is a big issue [in observational studies]. Everything we look at is very biased. Hard to get clarity on what data it’s trained on'' (P9).
What \toolname provides is to translate the HMM outputs into visual patterns.
P4 said ``The way it sets up helps to explain the model. It mitigates underlying criticism about the black-box nature of the model.''
P7 also commented ``You can kind of see the quality of data: sampling and visit intervals. That’s quite useful.''
\toolname is agnostic to model configurations (e.g., number of states) or data quality (e.g., sampling frequency) and lets users ask difficult questions to answer without visualizations: ``By looking at different models and data using \toolname, we can ask questions like, is there any genetic polymorphism why children have different trajectories?'' (P4).

\textbf{Multiple views of \toolname help users summarize, search patterns, and build subgroups of disease progression trajectories.}
Among many views, \pathwayfall and \heatmap were considered the most useful.
P2 also mentioned ``[\pathwayfall] catches my eyes immediately. It shows how people progress across different states over time.'' 
\pathwayfall shows the trends without hiding individuals (\emph{T2}), as participants noted ``It shows the impression of the volume of subjects. It shows how people travel across states over time in different ways. Summary (aggregated visualization) hide those'' (P3) and ``Instead of quantiles and distributions, you can see individuals'' (P5).
Participants find \heatmap useful to characterize the states discovered by HMMs.
P1 described the column of \heatmap as ``semantics of states'', which show what each state is (\emph{T1}).
\subjectlist were helpful to check the details: ``easy to understand, and there you can see huge variation among subjects'' (P7).
\kernel were useful to determine characteristics of cohorts (\emph{T3}): ``it helps you determine the samples' rapid progressors, slow progressors, or non progressors'' (P4).
To build and refine subgroups (\emph{T5}), \query was regarded helpful: ``I really love \query. I think it’s very difficult to query state sequence patterns from longitudinal data. Using \query, I can say anybody who had GADA and then IAA, I want to see those people.''

\textbf{It takes time to learn how to use \toolname.} Participants reported that it takes time and effort to learn how to use \toolname.
The major difficulty was understanding and learning how to use the multiple views.
P1 revealed, ``each view makes sense, but at first I did not know where I need to look to understand transitions and semantics of states.''
P6 also commented, ``Haven't seen this kind of visualization before, it is not easy to catch what it means at first [...]  it's not something that easily pops up, it's more like you spend time and effort.''
The difficulties come from not having a computational background and lack of experience with visual analytics applications.
P2 mentioned, ``some views can be difficult to understand, depending on his/her background.''
P5 said, ``Somebody who doesn't know HMMs need to learn a bit about HMMs first.''
P9 also added, ``It might be misleading for those who are not familiar with the algorithm part.''
However, all participants agreed that they were able to understand how to use \toolname after the series of workshops.
The key is to use \toolname multiple times repeatedly and to actively formulate and validate hypotheses on real data.
P7 mentioned ``It takes time and effort and many trials and errors. You have to be very eager to ask questions in order to understand them.''
P4 commented, ``The series of workshops and discussions really help me to better understand the tool. [It shows] the way one builds the analysis using the cohorts from the literature. I immediately grasp the way the data displayed.''

\textbf{Users want to provide feedback to the HMM learning process.} Participants shared ideas for future work.
Some participants wanted to add more \emph{automated analysis}.
P1 wanted \toolname to ``automatically identify and highlight outliers in \pathwayfall.''
Participants wanted to go beyond interpreting outcomes of trained HMM and to steer the model.
P4 wanted an ability ``to modify states and to establish new states.''
P9 also added ``Allow me to specify my own state. I would like to train a model with states I define. I would like to see how the dataset fits my mental model. It's a human-supervised model. I would like to have more control over the training part.''
Participants also want to add statistical tests to confirm the patterns they find in the visualization: ``Statistics will confirm the trends and help us build clinical trials. Tests like Chi-Square, T-test, [...] to test difference between subgroups in terms of progression rates and speeds'' (P4).
Some participants suggested the potential use of \toolname for \emph{clinician-patient interaction}.
P7 wanted to use \pathwayfall to explain a patient's progress: ``I’d like to have his/her waterfall with the background of others. My study subjects are at some point of the trajectory, so I’d like to use this to show where the subject is now. It’s more like a clinical tool. For patient education, I think it will be useful. For clinician education, it will also be useful.''
P4 said ``The utility of this tool could be for physicians, families, and others, who are interested in the particular person in relation to others. I think it will be a fantastic educational tool for clinicians, scientists, and families.''

\vspace{-.25cm}
\section{Discussion}
\label{sec:discussion}

In this section, we discuss some implications of \toolname, lessons learned from our study, and provide some thoughts for future work.

\vspace{-.25cm}
\subsection{HMMs and \toolname for Disease Progression Patterns}
\label{subsec:hmmva}

Clinical researchers want to understand why (etiology) and how (pathogenesis) a certain disease develops from longitudinal observational study data. 
HMMs are great ways to summarize the course of disease development. 
The states discovered by HMMs are explainable by associated variables (\emph{G1}) and naturally form trajectories that share the same state transition patterns (\emph{G2}).
However, as our participants revealed in Section~\ref{sec:user_feedback}, HMMs alone are not interpretable nor useful for clinicians to understand and explore the disease progression patterns effectively.
\toolname allows users to understand the state characteristics and disease progression patterns transparently through multiple, coordinated visualizations.
Furthermore, the interactive features of \toolname, such as \query, allow users to flexibly define subgroups and to test their relationships with various health-related outcomes (\emph{G3}).
As Section~\ref{sec:usage_scenario} describes, users can gain clinically meaningful insights by driving their analysis using \toolname.
The three clinical research goals (\emph{G1-3}), the five analytic tasks (\emph{T1-5}), and the design of \toolname can provide useful guidance for future visual analytics applications for clinical research.  
Our study can provide insights for other domains, where research questions involve summarizing and understanding temporal event sequence patterns.

There are recommendations and limitations when one uses HMMs and \toolname.
As mentioned in Section~\ref{sec:casestudy}, it is recommended to train an HMM with less than 20 states in order to maximize the interpretability.
Users can use any number of variables for running an HMM.
In some cases, users use latent variables, which are reduced from original variables using dimensionality reduction algorithms, for inferring states.
This can also reduce the interpretability of the model.
\pathwayfall can suffer from overplotting when the number of subjects and the number of visits increase.
It had no issues of overplotting for one of the largest observational study data from T1DI described in Section~\ref{sec:usage_scenario}.
In case of overplotting, one can adjust the canvas size and/or hide dots in \pathwayfall temporarily.
\heatmap could be overwhelming to users due to numbers written over bubbles, so it is recommended to only show the numbers when requested.

\vspace{-.25cm}
\subsection{Supporting Subgroup Building for Clinical VA}
\label{subsec:multifacet}

Our study demonstrates the usefulness of subgroup building activities (\emph{T5}) in clinical visual analytics.
As Section~\ref{sec:user_feedback} reports, users want to test hypotheses from prior literature and to conduct sensitivity analyses repeatedly by stratifying subjects into relevant subgroups.
Thus, it is very important for a visual analytics system to support creating and updating subgroups.
While designing \toolname, we discovered that every view can serve as a tool to create and refine subgroups of subjects based on users' constraints.
Since every view has its own perspective, users can express their constraints in different ways.
By combining multiple constraints, users can refine cohorts that meet their analysis goals.
For instance, a user can set a filter based on time of state transition using \query and another filter based on heights of subjects at a state using \detail.
The flexible methods of \toolname for constructing filters was useful for our experts to formulate and test their hypotheses.
The features blends well with users' analytic workflow, where they define and express cohorts and test their hypotheses by comparing the patterns of those different cohorts.

Supporting subgroup building activity can be adapted into future clinical visual analytics applications.
Clinical researchers often test hypotheses by comparing different subgroups.
The subgroups are frequently referred to with special names given by clinicians or previous literature, and they are the main character of stories they discuss in order to gain clinically meaningful insights. 
For instance, clinical researchers speculate whether the ``Multiple-AAB First'' group, discovered in Section~\ref{sec:usage_scenario}, can be replicated in other observational studies.
Future researchers can investigate methods to replicate subgroups from prior literature in an automated or human-in-the-loop approach, as requested by participants in Section~\ref{sec:user_feedback}.
In future work, we aim to investigate intuitive ways to build cohorts, which include drawing expected patterns over time-series visualizations like AxiSketcher~\cite{kwon17axisketcher}.

\vspace{-.25cm}
\subsection{Design Study Reflection: Working in a Triad}
\label{subsec:healthprof}

An important key to success of a design study is to effectively facilitate communication between experts in their domains, mainly health professionals, statisticians, and visualization experts.
By combining the pieces from the triad of experts, we can build an effective visual analytics application. 
Though it is essential for the experts to carefully evaluate each other's inputs,
it is very challenging due to the knowledge gaps between domains. 
What facilitates effective communication between experts in different fields can be the visualization itself~\cite{kwon2016visohc}.
Throughout meetings, experts discussed over \toolname, which resulted in several changes to try out in the next phase. 
Therefore, what makes a visual analytics tool interpretable and engaging for users is how effectively the representations facilitate communication and new insights among the team.
\toolname was used to translate a fundamental question from health professionals about HMMs, ``what is a state and how does it differ from clinical stages?''
As Section~\ref{sec:casestudy} revealed, \toolname allowed our team to understand how multiple states may fit a clinical stage and to hypothesize more granular states that can describe within and outside clinical stages.
An expert also reported that \pathwayfall allowed him to make sense of the overall scale and dimension of patients' visits over time. 
In the meetings, we gained numerous insights into the data aided by the various visualizations presented in this paper.
In the end, the design process takes long but mutual learning across domains as Hall et al.~\cite{hall_design_2019} nicely depict.

\vspace{-.25cm}
\subsection{Learning Curve, Uncertainties, and Trust}
\label{subsec:onboard}

We find that clinicians often encounter problems while understanding layered, unknown uncertainties in the visual analytics pipeline.
First, users need to understand how multiple views are set up.
Second, they need to understand what the underlying model conveys in terms of its inputs and outputs.
Third, they also need to know how each original study was designed and how the subjects were sampled and measured.
Depending on the background of users, they might fall into any of the three problems, lose trust in the visualizations, and/or make false interpretation of the results, as Sacha et al. provides~\cite{sacha_role_2016}.
We overcome the issues through a series of workshops. 
Participants reported that the key to overcoming such problems was to repeatedly formulate and validate hypotheses with small examples from real data.
Participants often expect insights to pop out without any actions.
It is important for them to keep in mind that the clinicians need to drive the analysis by actively formulating and validating hypotheses. 
In addition, designers need to become semi-experts in clinical and analytic domains by actively engaging with the end users for a long period of time.
We believe that clinicians start engaging with the tool when they are convinced by useful findings, insights, and hypotheses proven through the tool.
Thus, designers of the system need to spend time in understanding how the model works and interpreting the results in clinical context.
It is often very difficult to acquire additional skills in machine learning and clinical domains.
We believe that tight collaboration through a study group will be useful to learn the perspectives of clinicians.
The design study protocol clearly specifies the role of liaison, which we need to pursue in order to properly implement the visual analytics approach for medical experts~\cite{simon2015bridging}.

\vspace{-.35cm}
\section{Conclusion}
\label{sec:conclusion}

In this work, we conducted a design study with clinical researchers, including physician scientists, statisticians, and visualization experts, to develop a visual analytics application for exploring disease progression pathways and their interaction with various measures.
As a result, we developed \toolname, which seamlessly integrates HMMs and provides views and interactive features that facilitate users to formulate and test hypotheses by iteratively constructing multiple subgroups based on state transitions and distributions over multiple measures and then analyzing and comparing them.
The usage scenario and user experiences sections demonstrate the usefulness of the application to gain a summary of disease progression trajectories transparently, to construct subgroups in a flexible manner, and to characterize the patterns using relevant health outcomes.
Our collaborators have been continuously researching disease progression patterns for type 1 diabetes, Huntington's disease, and Parkinson's disease using \toolname.
In future work, we plan to extend \toolname so that users can visually supervise the training process of HMMs.
We also plan to apply \toolname into clinical studies of other diseases (e.g., Alzheimer's Disease).

\vspace{-.35cm}
\section*{acknowledgements}

We want to thank all reviewers who provided thoughtful feedback throughout the revision process.
We also wish to thank the T1DI Study Group for their help in this work.
The T1DI Study Group consists of following members: 1) JDRF--Jessica Dunne, Olivia Lou; 2) IBM--Vibha Anand, Mohamed Ghalwash, Eileen Koski, Bum Chul Kwon, Ying Li, Zhiguo Li, Bin Liu, Ashwani Malhotra, Kenney Ng; 3) DiPiS--Helena Elding Larsson, Josefine J\"onsson, \AA{}ke Lernmark, Markus Lundgren, Marlena Maziarz; 4) BABYDIAB--Peter Achenbach, Christiane Winkler, Anette Ziegler; 5) DIPP--Heikki Hy\"oty, Jorma Ilonen, Mikael Knip, Jorma Toppari, Riitta Veijola; 6) DEW-IT--Bill Hagopian, Michael Killian, Darius Schneider; 7) DAISY--Brigitte Frohnert, Jill Norris, Marian Rewers, Andrea Steck, Kathleen Waugh, Liping Yu.
This work was supported in part by JDRF (1-IND-2019-717-I-X, 1-SRA-2019-722-I-X, 1-SRA-2019-723-I-X, 1-SRA-2019-719-I-X, 1-SRA-2019-721-I-X, 1-SRA-2019-720-I-X).

\newpage

\bibliographystyle{abbrv-doi}

\bibliography{DPVis}

\begin{IEEEbiographynophoto}{Bum Chul Kwon}
is Research Staff Member at IBM Research. His research area includes visual analytics, data visualization, human-computer interaction, healthcare, and machine learning. Prior to joining IBM Research, he worked as postdoctoral researcher at University of Konstanz, Germany. He received his M.S. and Ph.D. from Purdue University in 2010 and 2013, respectively. He received his B.S. in Systems Engineering from University of Virginia in 2008.
\end{IEEEbiographynophoto}

\begin{IEEEbiographynophoto}{Vibha Anand}
is a Research Staff Member at IBM Research. Her research focuses on applying machine learning techniques to prior observational studies, clinical trials, EMR data and real-time biosensors to derive actionable insights. Before IBM Research, she was an Assistant Professor and Professional Staff at the Cleveland Clinic and at Indiana University, School of Medicine in Dept. of Pediatrics. She holds a BE in Electrical Engineering, MS in Computer Science, PhD in Informatics and is a Fellow of AMIA.  
\end{IEEEbiographynophoto}

\begin{IEEEbiographynophoto}{Kristen A. Severson}
is a post-doctoral researcher at IBM Research. Her research focuses on the development and application of machine learning techniques to derive insight from real-world health data. Prior to joining IBM, she received a PhD from the Massachusetts Institute of Technology. She also holds a BS in chemical engineering from Carnegie Mellon University.
\end{IEEEbiographynophoto}

\begin{IEEEbiographynophoto}{Soumya Ghosh}
is a Research Staff Member at IBM Research. His research focuses on the design of flexible statistical models and efficient inference algorithms for reasoning about noisy, high-dimensional data. His recent work has examined approaches for combining the complementary strengths of Bayesian methods, graphical models, and deep neural networks for modeling progression of neuro-degenerative diseases. He holds a PhD in computer science from Brown University. 
\end{IEEEbiographynophoto}

\begin{IEEEbiographynophoto}{Zhaonan Sun}
is a Research Staff Member at IBM Research. Her research focuses on developing statistical and machine learning methods in the healthcare domain for generating insights from real-world data. Prior to joining IBM, she received PhD in statistics from Purdue University.
\end{IEEEbiographynophoto}

\begin{IEEEbiographynophoto}{Markus Lundgren}
is a M.D, Ph.D. specializing in pediatric endocrinology and a researcher at the Department of clinical sciences Malm\"o at Lund University, Lund, Sweden. His research focuses on pediatric type 1 diabetes Prediction and Prevention including the effects of early disease detection. He received his Ph.D. and M.D. from Lund University in 2017 and 2003, respectively.
\end{IEEEbiographynophoto}

\begin{IEEEbiographynophoto}{Brigitte I. Frohnert}
is a M.D., Ph.D. who specializes in a pediatric endocrinology in University of Colorado Denver, Aurora, Colorado. Her research interest is in understanding the environmental factors which contribute to the development of islet autoimmunity and progression to type 1 diabetes, with the ultimate goal of preventing type 1 diabetes in the future. She received her M.D and Ph.D from University of Minnesota in 2002 and 2000, respectively.
\end{IEEEbiographynophoto}

\begin{IEEEbiographynophoto}{Kenney Ng}
is a Principal Research Staff Member and manager of the Health Analytics Research Group at IBM Research. His research focus is on the development and application of AI techniques to analyze, model and derive actionable insights from real world health data. He received B.S., M.S., and Ph.D. degrees in electrical engineering and computer science from the Massachusetts Institute of Technology. He is a member of IEEE and AMIA.
\end{IEEEbiographynophoto}

\vfill

\end{document}